\pdfoutput=1
\documentclass[11pt]{article}
\usepackage{acl}

\usepackage[T1]{fontenc}
\usepackage[utf8]{inputenc}

\usepackage{times}
\usepackage{latexsym}
\usepackage{booktabs}
\usepackage{natbib}
\usepackage{amsmath,amssymb,amsfonts}
\usepackage{colortbl}
\usepackage{graphicx}
\usepackage{tabularx}

\usepackage{listings}
\usepackage{pgfplots}
\usepackage{pgfplotstable}
\pgfplotsset{compat=1.18}
\usepackage{multirow}
\usepackage{rotating}
\usepackage{tablefootnote}
\usepackage{longtable}
\usepackage{makecell}

\usepackage{enumerate}
\usepackage{lingmacros}
\usepackage{fancyvrb}
\usepackage{tikz}
\usetikzlibrary{positioning,arrows.meta,fit}
\usepackage[dvipsnames]{xcolor}

\usepackage{pifont}
\usepackage{appendix}
\usepackage{xspace}
\usepackage{microtype}
\usepackage{inconsolata}
\usepackage{placeins}

\definecolor{darkpastelgreen}{rgb}{0.01, 0.75, 0.24}
\definecolor{darkpastelred}{rgb}{0.76, 0.23, 0.13}

\makeatletter
\renewcommand\paragraph{\@startsection{paragraph}{4}{\z@}%
  {0.5ex \@plus 0.2ex \@minus 0.2ex}%
  {-1em}%
  {\normalfont\normalsize\bfseries}}
\makeatother

\newcommand{\method}{\textsc{TermJudge}\xspace}
\newcommand{\Conforme}{\textsc{Conforming}\xspace}
\newcommand{\VarJ}{\textsc{Justified}\xspace}
\newcommand{\VarNJ}{\textsc{Unjustified}\xspace}

\newcommand{\Aone}{\textbf{A1}\xspace}
\newcommand{\Atwo}{\textbf{A2}\xspace}
\newcommand{\Athree}{\textbf{A3}\xspace}
\newcommand{\Afour}{\textbf{A4}\xspace}
\newcommand{\Afive}{\textbf{A5}\xspace}
\newcommand{\Asix}{\textbf{A6}\xspace}
\newcommand{\Aseven}{\textbf{A7}\xspace}

\newcommand{\biomqm}{bio-MQM\xspace}
\newcommand{\biomqmterms}{BioMQM-Terms\xspace}

\newcommand{\acceq}{acc$^{*}_{\text{eq}}$\xspace}
\newcommand{\STEP}{\textsc{Step}\xspace}
\newcommand{\ACL}{\textsc{ParaNLP}\xspace}
\newcommand{\iwslt}{\textsc{IWSLT23}\xspace}
\newcommand{\Concordancer}{\texorpdfstring{\textsc{Concordancer}}{Concordancer}\xspace}

\title{\method: A Document-Level Metric Judging, Not Counting, Terminology in Machine Translation Evaluation}

\author{Nicolas Dahan$^{\diamondsuit\spadesuit}$ 
\quad François Yvon$^\spadesuit$ \quad Rachel Bawden$^\diamondsuit$\\
$^\diamondsuit$Inria Paris, France \\
$^\spadesuit$Sorbonne Université, CNRS, ISIR, Paris, France \\
\texttt{\{nicolas.dahan,rachel.bawden\}@inria.fr} \quad \texttt{yvon@isir.upmc.fr}
}
\begin{document}
\maketitle

\begin{abstract}
Existing automatic metrics for evaluating terminological use in machine translation (MT) penalise any divergence from a fixed reference, conflating translation errors with the valid terminological variation that human translators routinely produce. We introduce \method, a document-level terminology metric that assigns an interpretable verdict to every term occurrence: glossary-conforming occurrences are settled deterministically, while divergences are assessed under a two-step LLM-as-a-judge procedure using the full document context: the first detects and labels terminology errors; the second sorts valid document-level variations from inconsistencies.
Validated against expert error annotations and document-level human MQM scores, \method\ ranks first in both system- and segment-level meta-evaluation, ahead of glossary-conformity and quality-estimation baselines. When applied to eight systems translating academic documents, under two prompting conditions, we observe that glossary injection improves terminology translation in all paired comparisons, by removing genuine errors rather than valid variation.
\method\ is released as open-source code.\footnote{\url{https://github.com/nicolasdahan/termjudge}}

\end{abstract}

\section{Introduction}
\label{sec:introduction}

Evaluating machine translation (MT) at the document level in specialised settings requires assessing how systems translate the technical vocabulary of those documents. In most documents, technical concepts can be split into two main groups: rare concepts that only occur once or twice, usually under the same surface form; and frequent concepts, which may appear under multiple guises. For example, in a research paper on MT, a concept such as \texttt{machine translation} may occur repeatedly, as the preferred form (\textsl{machine translation}), an acronym (\emph{MT}), a reduction (\emph{translation}), a lexical substitution (\emph{automatic translation}), or a morphosyntactic reformulation (\emph{machine translated (text)}) \citep{daille:hal-01693035}; translators may amplify this variation and introduce further alternations for stylistic or discourse reasons \citep{bowker-1998,fernandez-silva-kerremans-2011}.

Existing automatic metrics for evaluating the translation of terms decide whether a target form is acceptable through \emph{matching}: against a term-annotated reference \citep{farajian-etal-2018-evaluation}, against a bilingual glossary entry \citep{alam2021evaluationmachinetranslationterminology,semenov-etal-2025-findings}, against the document's other occurrences of the same concept \citep{itagaki-etal-2007-automatic,semenov-bojar-2022-automated}, or against source-side variation patterns \citep{dahan-etal-2026-improving}. None of them asks whether a  mismatch is a translation error or a valid translation choice (e.g.,~target-language amplification, polysemy disambiguation, acronym introduction at first mention, stylistic alternation), so they may conflate acceptable translations with errors. General-purpose neural metrics offer no remedy, losing correlation with human judgement precisely on the specialised domains where terminology errors concentrate \citep{zouhar-etal-2024-fine}. The question we address is therefore whether an automatic metric can decide, for each occurrence of a concept, whether the output form is acceptable rather than merely identical to a predetermined reference.
We propose to address this gap with \method, an automatic document-level terminology metric that not only counts \emph{divergences} (translations that depart from the expected form of a term) but judges each one, and returns interpretable verdicts alongside an aggregate score. Our contributions are:
\begin{itemize}
    \itemsep0pt
    \item A concept-aware document-level terminology metric, built on an LLM-as-a-judge approach comprising several steps (reference validation, then error classification and document-consistency qualification), grounded in established translation studies, which discriminates errors from valid variation and produces an \emph{interpretable verdict profile} for each system rather than a single opaque score.
    \item \biomqmterms, a terminology-evaluation resource derived from \biomqm \citep{zouhar-etal-2024-fine}, a corpus containing MT outputs in the biomedical domain, with error annotations. To this we add a bilingual glossary of 729 English--French biomedical concepts and 13{,}200 aligned term--translation pairs for ten MT systems, released with our code.
    \item A series of empirical validations: per-occurrence agreement with a fine-grained human error typology on a geoscience corpus (\STEP; \citealp{carcamo:hal-05560658}), a document-level meta-evaluation on \biomqmterms, where \method\ ranks first under every aggregation of its score, ahead of ten alternative metrics, and a reevaluation of the use of glossaries to guide LLM-based MT: on two NLP test corpora, we find that glossary-guided prompting improves term translation.
\end{itemize}

\section{Related Work}
\label{sec:related}
For the global evaluation of MT quality, some of the most widely used automatic metrics (alongside traditional surface-based metrics such as BLEU \citep{papineni-etal-2002-bleu} and chrF \citep{popovic-2015-chrf}) are fine-tuned neural models such as \textsc{Comet} \citep{rei-etal-2020-comet} and \textsc{MetricX} \citep{juraska-etal-2024-metricx}, together with their reference-free counterparts \citep{rei-etal-2022-cometkiwi}. They are trained and mostly used at the sentence level. Despite proposals to extend them to longer spans \citep{vernikos-etal-2022-embarrassingly,deutsch-etal-2023-training}, their document-level application remains questionable \citep{dahan:hal-05663019}. Two further properties limit their use for terminology. First, fine-tuned metrics lose substantial correlation with human judgements on unseen specialised domains, precisely where terminology errors concentrate \citep{zouhar-etal-2024-fine}. Second, such metrics summarise all aspects of translation quality in a single scalar score, failing to distinguish terminology errors from other types of mistranslations.

Evaluation practice is therefore increasingly dominated by LLM-as-a-judge approaches. \textsc{Gemba} involves eliciting direct quality scores \citep{kocmi-federmann-2023-large}, and \textsc{Gemba-MQM} annotating error spans within the generic MQM typology \citep{burchardt-2013-multidimensional,kocmi-federmann-2023-gemba}; \citet{minder-etal-2025-testing} test how reliably LLMs annotate specialised translations with such error typologies: error detection reaches satisfactory levels, but accurate categorisation requires injecting the error definitions in the prompt. LLM-as-a-judge approaches bring the flexibility that qualitative judgements require, but they carry documented biases (order of presentation, verbosity preference, instability across rephrased queries; \citealp{zheng2023judgingllmasajudgemtbenchchatbot}), making careful meta-evaluation against human judgements indispensable \citep{mathur-etal-2020-tangled,deutsch-etal-2023-ties}. Whether applied to isolated segments or to whole documents, they annotate general error categories and do not track how consistently a concept is rendered across a document.

Dedicated terminology evaluation has mostly relied on matching the output against expected target forms. Given a term-annotated reference translation, \citet{farajian-etal-2018-evaluation} compute a term hit rate, a BLEU-like clipped count of the reference terms retrieved in the output, which TermEval \citep{haque-etal-2023-evaluating} extends by also accepting the lexical and inflectional variations of each reference term listed in its termbank. Given a bilingual glossary entry, metrics count how often the expected target term appears in the output, from exact surface matching \citep{alam2021evaluationmachinetranslationterminology} to lemma-aware matching \citep{alam-etal-2021-findings}, as adopted in the WMT terminology shared tasks \citep{alam-etal-2021-findings,semenov-etal-2023-findings}, whose 2025 edition extends the exercise to document-level translation with one-to-many dictionaries \citep{semenov-etal-2025-findings}.
Whatever the matching strategy, these metrics remain anchored to an inventory of expected forms: unlisted reductions, context-dependent acronyms and lexical variants are counted as failures irrespective of their adequacy in context.

When no glossary is available, consistency metrics score the target side against itself. \citet{semenov-bojar-2022-automated} match each occurrence against a pseudo-reference derived from the system's own dominant translation, concentration indices such as the Herfindahl--Hirschman index \citep{itagaki-etal-2007-automatic,Gapar2022MeasuringTC} and the Lexical Translation Consistency Ratio \citep{lyu-etal-2021-encouraging,wang-etal-2025-delta} reward the reuse of a single form, and cross-term coherence extends the family to the source side, counting how many variation relationships survive translation \citep{dahan-etal-2026-improving}. All share one assumption: a term is well translated when every occurrence receives the same translation. That assumption keeps them cheap and reference-free, but blind to the difference between undue changes and motivated alternations. It is also only warranted for prescriptive terminologies: terminology standards themselves recognise a scale of acceptability from preferred to merely admitted forms \citep{iso1087-2019}.

That difference is precisely what studies of specialised discourse document. Denominative variation is a constitutive property of terminology in running text \citep{daille:hal-01693035}, refined into graphical, morphosyntactic, reduction, expansion and lexical variants by \citet{carcamo:hal-05442584}. Translators preserve or even accentuate source-side variation \citep{fernandez-silva-kerremans-2011}, alternate between several translations for discourse and stylistic reasons \citep{vinay-darbelnet-1972,bowker-1998,pecman:hal-01232653}, introduce pragmatic and explicitness shifts \citep{chesterman-2016,blumkulka1986shifts}, and resist the over-standardisation that erases author-intended distinctions \citep{bowker-hawkins-2006}. MT systems, by contrast, produce less variation than humans \citep{culo-nitzke-2016-patterns}. These findings have not reached automatic evaluation: no existing terminology metric asks whether a divergence from the expected form is one of these motivated variations or an error.

\section{The \method{} Metric}
\label{sec:metric}
\begin{figure}[t]
\centering
\definecolor{vtInk}{HTML}{263238}%
\definecolor{vtTeal}{HTML}{00796B}%
\definecolor{vtTealL}{HTML}{B2DFDB}%
\definecolor{vtOk}{HTML}{2E7D32}%
\definecolor{vtOkBg}{HTML}{E8F5E9}%
\definecolor{vtBad}{HTML}{C62828}%
\definecolor{vtBadBg}{HTML}{FFEBEE}%
\begin{tikzpicture}[
    font=\footnotesize, every node/.style={align=center},
    dot/.style={circle, fill=vtTeal, text=white, font=\scriptsize\bfseries, minimum size=0.52cm, inner sep=0pt},
    card/.style={fill=black!6, rounded corners=3pt, text=vtInk, inner xsep=6pt, inner ysep=3pt, minimum width=3.45cm},
    okpill/.style={draw=vtOk, fill=vtOkBg, text=vtOk, rounded corners=5pt, font=\scriptsize\bfseries,
                   inner xsep=5pt, inner ysep=2.5pt, anchor=west},
    badpill/.style={draw=vtBad, fill=vtBadBg, text=vtBad, rounded corners=5pt, font=\scriptsize\bfseries,
                    inner xsep=5pt, inner ysep=2.5pt, anchor=west},
    rail/.style={draw=vtTealL, line width=2.5pt, line cap=round},
    go/.style={-{Stealth[length=4pt, width=3pt]}, line width=0.6pt, draw=vtTeal},
    lbl/.style={font=\scriptsize\itshape, text=black!65, inner sep=1.5pt}
]
\def\xr{0.40}\def\xc{2.60}\def\xo{5.25}
\def\yA{4.10}\def\yB{2.80}\def\yC{1.50}
% rail d'etapes (index)
\draw[rail] (\xr,\yA) -- (\xr,\yC);
\node[dot] at (\xr,\yA) {1--2};
\node[dot] at (\xr,\yB) {3};
\node[dot] at (\xr,\yC) {4};
% cartouches : nom d'etape + question
\node[card] (c1) at (\xc,\yA) {{\fontsize{6.5}{7.5}\selectfont\color{black!55}\textsc{steps 1--2} \textperiodcentered{} reference \& matching}\\[1pt]$t = R(s)$?};
\node[card] (c2) at (\xc,\yB) {{\fontsize{6.5}{7.5}\selectfont\color{black!55}\textsc{step 3} \textperiodcentered{} error judge}\\[1pt]Q1: error?};
\node[card] (c3) at (\xc,\yC) {{\fontsize{6.5}{7.5}\selectfont\color{black!55}\textsc{step 4} \textperiodcentered{} consistency judge}\\[1pt]Q2: justified?};
% verdicts
\node[okpill]  (conf)  at (\xo,\yA)      {\ding{51}\; \Conforme};
\node[badpill] (err)   at (\xo,\yB)      {\ding{55}\; A1..A7, C..H};
\node[okpill]  (varj)  at (\xo,\yC+0.33) {\ding{51}\; \VarJ};
\node[badpill] (varnj) at (\xo,\yC-0.33) {\ding{55}\; \VarNJ};
% entree
\draw[go] (\xc,4.98) -- (c1.north);
\node[lbl, anchor=west] at (\xc+0.06,4.80) {aligned occurrence $(s \to t)$};
% etapes 1-2 et 3 : a droite vers le verdict, en bas vers la suite
\draw[go]    (c1.east)  -- node[lbl, above] {match} (conf.west);
\draw[go]    (c1.south) -- node[lbl, right] {divergence} (c2.north);
\draw[go]    (c2.east)  -- node[lbl, above] {error} (err.west);
\draw[go]    (c2.south) -- node[lbl, right] {no error} (c3.north);
% etape 4 : une fleche qui se divise vers les deux verdicts finaux
\path (c3.east) ++(0.38,0) coordinate (s4);
\draw[go] (c3.east) -- (s4) |- (varj.west);
\draw[go] (s4) |- (varnj.west);
\path (s4 |- varj.west) -- node[lbl, above] {yes} (varj.west);
\path (s4 |- varnj.west) -- node[lbl, below] {no} (varnj.west);
\end{tikzpicture}
\caption{The \method\ decision flow (\S\ref{ssec:pipeline}): a match with $R(s)$ is considered \Conforme\ without any LLM call; a divergence is classified by the error judge (Q1) then, if error-free, by the consistency judge (Q2), as \VarJ\ or \VarNJ.}
\label{fig:verdict-tree}
\end{figure}
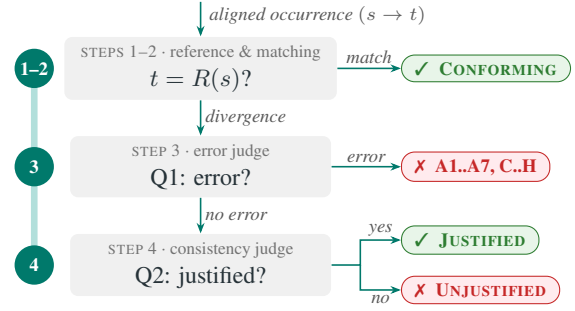

\method evaluates the terminological adequacy and consistency of a translated document by assigning a verdict to the translation of every term occurrence, and by aggregating these verdicts into a document-level score that can rank systems on terminological quality. A preprocessing stage turns a terminology and a translated corpus into aligned term occurrences (\S\ref{ssec:data-requirements}); \method\ itself is a five-step pipeline (\S\ref{ssec:pipeline}) whose decision flow is displayed in Figure~\ref{fig:verdict-tree}. Appendix~\ref{app:pipeline} (Figure~\ref{fig:pipeline}) gives a graphical overview of both stages.
\method\ shares its starting point with the consistency tradition (\S\ref{sec:related}): each source form is \emph{evaluated separately}, through the set of translations it receives across the document. The variation being adjudicated is therefore target-side only, source-side variation being factored out by construction. 
For this, \method\ first computes a reference translation for each source form, and flags as \emph{divergence} any translation that departs from it. It then submits each divergence to an LLM judge in two steps: a first question, Q1, asks whether the divergence is an error and of which kind. When no error is found, a second question, Q2, asks whether the variation is a valid choice justified by its context. Both questions are answered by the same model using different prompts, called the \emph{error judge} and the \emph{consistency judge} in what follows. The decomposition is deliberate. The two questions do not judge the same object: Q1 assesses the translation itself, so its answer is independent of the occurrence's position and can be shared across identical translations (\S\ref{ssec:aggregation}), and Q2 assesses the consistency of a choice with the concept's other translations in the document, and is asked per occurrence. This reflects a key distinction in our error typology: between errors in a translation taken on its own and inconsistencies between the translations of a concept within a document (Appendix~\ref{app:typology}). Merging the two questions into a single LLM call would make every judgement positional, at the occurrence level. It would also imply a much larger label set, possibly degrading the judge's decisions (Appendix~\ref{app:typology-comparison}). Answering Q1 at the form-level does not mean judging it in isolation: like for Q2, the judge also receives the corresponding concept's glossary entry and the translations of its other source forms.

\subsection{Data Requirements and Preprocessing}
\label{ssec:data-requirements}
\method\ relies on a single external resource, a bilingual terminology of the domain structured as a SKOS glossary:\footnote{Simple Knowledge Organization System, \url{https://www.w3.org/TR/skos-reference/}.} each entry pairs a \emph{concept} with one \emph{preferred term} (or \emph{head term}) per language, and a list of \emph{alternative terms}, or \emph{variants}.
The object of the evaluation is a collection of source documents together with their translations by one or more MT systems, with aligned source and target segments; no human reference translation is needed.

A glossary, however large, cannot delimit the forms under which concepts surface: beyond \emph{glossary variants} listed as alternative terms, running texts also introduce \emph{non-glossary variants} (reorderings, reductions, context-dependent acronyms). 
Term occurrences are therefore identified with the \Concordancer,\footnote{\url{https://gitlab.inria.fr/almanach/concordancer}} an in-house tool that takes the glossary as input and detects, in each source document, the occurrences of every concept. This raises coverage beyond exact-form and lemma matching by also recognising variants derived from glossary entries or detected in context: on \ACL\ and \iwslt, lemma matching alone would miss about 15\% of the occurrences the \Concordancer\ detects \citep{dahan-etal-2026-improving}.

Each attested source form is then labelled automatically with its variation category relative to the head term (procedure and validation in Appendix~\ref{app:pipeline}), following the typology of \S\ref{sec:related}: \emph{no variation}, \emph{graphical}, \emph{morphosyntactic}, \emph{reduction}, \emph{expansion}, \emph{lexical} or \emph{combined} \citep{carcamo:hal-05442584} (one example per category in Table~\ref{tab:variation-examples}, Appendix~\ref{app:pipeline}). Each occurrence is then aligned with the target-side span that corresponds to its translation, identified by an LLM call within the aligned target segment rather than the whole document;\footnote{\texttt{gpt-4.1-mini}; details in Appendix~\ref{app:pipeline}.} both sides are normalised for case, punctuation and whitespace, and lemmatised identically.

\method\ starts from the concept and isolates every distinct source form attested in the document; the head term itself need not appear. The set of occurrences of one source form is a \emph{chain}, and a chain of size one is a \emph{hapax}. Within a chain, occurrences share the source form and differ by their position in the document, hence by their context, and by the translation the system produced for each of them.
\method\ returns one verdict per occurrence (Table~\ref{tab:verdicts}); Appendix~\ref{app:pipeline} (Tables~\ref{tab:worked-example} and~\ref{tab:example-chain}) illustrates the setting on occurrences from real evaluation runs.

\begin{table}[t]
\footnotesize
\centering
\setlength{\tabcolsep}{4pt}
\renewcommand{\arraystretch}{0.95}
\resizebox{\columnwidth}{!}{%
\begin{tabular}{@{}llc@{}}
\toprule
\textbf{Verdict} & \textbf{Description} & \textbf{Penalty} \\
\midrule
\Conforme & target matches $R(s)$ (Step~1), or judged error-free & 0 \\
\VarJ     & variation, consistent with document usage & 0 \\
\VarNJ    & variation, inconsistent (incl.\ neutralisation) & 0.5 \\
\midrule
\Aone     & attested term of a neighbouring concept   & 1 \\
\Atwo     & paraphrase instead of term                & 1 \\
\Athree   & invented literal calque                   & 3 \\
\Afour    & left untranslated                         & 4 \\
\Afive    & do-not-translate item translated          & 4 \\
\Asix     & wrong internal constituent                & 4 \\
\Aseven   & general-language word instead of term     & 3 \\
\textbf{C} & mis-parsed term structure                & 4 \\
\textbf{D} & wrong semantic relations between constituents & 3 \\
\textbf{E} & failed transposition / syntactic calque  & 1 \\
\textbf{F} & grammar / phraseology around the term    & 1 \\
\textbf{G} & content altered (addition, omission, hallucination) & 1 \\
\textbf{H} & defective target-language expression / register & 1 \\
\bottomrule
\end{tabular}%
}
\caption{The sixteen verdicts of \method\ and their penalties. \Aone--\Aseven\ follow level~A of the error typology \citep{carcamo:hal-05560658}; \textbf{C}--\textbf{H} group its remaining families, each scored at its most lenient sub-code (benefit of the doubt). Appendix~\ref{app:penalties} explains the weights; Appendix~\ref{app:typology} gives the full typology.}
\label{tab:verdicts}
\end{table}

\subsection{Evaluation Pipeline}
\label{ssec:pipeline}

\paragraph{Step 1: Reference selection.}
\label{ssec:reference-selection}
For each source form $s$ of a concept in a document, \method\ fixes the expected reference translation $R(s)$: the glossary translation when the source form has one, considered as authoritative, and otherwise the translation that the evaluated system itself produces most often in the document (on lemmas, ties broken by first appearance), a \emph{pseudo-reference} generalising those of consistency metrics \citep{semenov-bojar-2022-automated}. A pseudo-reference is first validated by the error judge of Step~3 on source-side evidence only;\footnote{Being borrowed from the very output under evaluation, it cannot be taken at face value; the circularity is made explicit to the judges.} if judged erroneous, it is replaced by a reference generated by an LLM from the concept entry and source-side contexts, never from the systems' outputs. Each reference is recorded with its provenance (glossary, validated or generated).

\paragraph{Step 2: Divergence detection.}
\label{ssec:matching}
Each translation is compared with $R(s)$ by exact string matching on lemmas, ignoring case, articles, hyphens and diacritics. A match is labelled \Conforme\ without any LLM call: the metric evaluates the choice of term, not its written form. A mismatch is a \emph{divergence} in the sense defined above, and in this case we proceed to Step~3.

\paragraph{Step 3: Q1, error judgement.}
\label{ssec:q1}

Each divergence's correctness is evaluated by an LLM call that receives the aligned pair, the current segment, the concept's glossary entry when available, the preceding aligned segments, and the concept's prior translations grouped by source form. The question is \emph{absolute}: is the target a valid, contextually-appropriate translation of the source term? The candidate error labels come from a simplified version of a typology developed for the human annotation of terminology errors in specialised MT \citep{carcamo:hal-05560658}, with a severity scale following \citet{benard-etal-2024-etude}: wrong equivalent selection uses fine-grained labels (\Aone--\Aseven), the six other error families are collapsed into one code each (\textbf{C}--\textbf{H}, Table~\ref{tab:verdicts}), and the in-document inconsistency family (B) is excluded, being exactly the question delegated to the consistency judge of Step~4 (details in Appendix~\ref{app:typology}). The judge answers with \texttt{no\_error} or one of the 13~labels, whose gravity fixes the penalty. An error label is the occurrence's final verdict, while \texttt{no\_error} routes the occurrence to Step~4. Hapax forms skip the matching shortcut of Step~2 and are always submitted to Q1: when a form occurs once, its pseudo-reference is derived from that very occurrence, so a match would be circular. Since consistency is undefined for a single occurrence, an error-free hapax is directly labelled \Conforme.

\paragraph{Step 4: Q2, consistency judgement.}
\label{ssec:q2}

Divergences judged error-free by Q1 are passed to the consistency judge, which decides whether the translation is coherent given how the concept is rendered elsewhere in the document. It sees the concept's translations grouped by source form, with the dominant translation and first mentions marked. It answers with one of ten labels, grounded in terminology and translation studies \citep{vinay-darbelnet-1972,bowker-1998,pecman:hal-01232653} as detailed in Appendix~\ref{app:q2-labels}. Seven name a reason that makes the variation coherent (`first mention', `explicitation', `avoid repetition', `document usage', `register', `facet', `synonym merge') and yield \VarJ. Three name an inconsistency and yield \VarNJ: `synonym inconsistency' and `gratuitous divergence' when one source form receives divergent translations, `neutralisation' when distinct source forms collapse onto one translation; together they operationalise family~B of the typology, cases combining an incorrect translation with instability being caught upstream by Q1.
Neutralisation, the reproduction of the translation established for \emph{another} source form of the same concept, erases a distinction made by the author \citep{bowker-hawkins-2006}, for example the case of a document alternating \emph{percutaneous nephrolithotomy} and its acronym \emph{PNL} where both are rendered as \emph{néphrolithotomie percutanée}.

\paragraph{Step 5: Propagation and scoring.}
\label{ssec:aggregation}

Q1 is called once for each distinct divergent translation of a source form within a document, and its verdict is propagated to the group's other occurrences; keeping the translation unlemmatised in the grouping key preserves number and gender errors. Q2, whose answer depends on the occurrence's position, is called once per occurrence of a correct form. Together with the matching shortcut of Step~2, this propagation divides the number of LLM calls by 2.55 on \iwslt\ while leaving the system ranking nearly unchanged (25 of 28 pairwise orderings; Appendix~\ref{app:exhaustive}).
Each verdict maps to a penalty (Table~\ref{tab:verdicts}), and the score of a document for a system is the mean penalty over all its scored occurrences, including those labelled \Conforme\ or \VarJ\ (penalty 0); lower is better, and scores are negated when correlated with human quality scores. System scores macro-average document scores, occurrences can equally be macro-averaged per source form or per concept, and the meta-evaluation of \S\ref{ssec:evaluation-methods} reports all three aggregations.

\subsection{Metric Configuration}
\label{ssec:default-config}

All experiments use a single configuration. The judges run on \texttt{gpt-4.1-mini}, chosen for its low cost (Appendix~\ref{app:cost}) and because it belongs to a different family from every evaluated system, which limits the self-preference bias of LLM judges \citep{zheng2023judgingllmasajudgemtbenchchatbot,panickssery2024llmevaluatorsrecognizefavor}; its judgements are validated against human annotations in \S\ref{sec:results}. References are selected as in Step~1; the prompts, their context and the decoding settings are detailed in Appendix~\ref{app:prompts}.

\section{Experimental Setup}
\label{sec:setup}

\subsection{Datasets}
\label{sec:datasets}

We use two human-annotated corpora to validate \method at two complementary granularities: \STEP\ \citep{carcamo:hal-05560658}, in the geoscience domain, provides per-occurrence gold error labels, enabling us to test whether judges classify individual occurrences as human annotators would. \biomqm\ \citep{zouhar-etal-2024-fine}, which provides document-level MQM annotations for ten systems, allows us to test how well aggregated document scores match human quality assessments. Two scientific NLP corpora, \ACL\ \citep{peng-etal-2026-parallel} and \iwslt\ \citep{salesky-etal-2023-evaluating}, serve as additional test beds on which we rank contemporary LLM-based MT systems. Each corpus is paired with a terminology (\S\ref{ssec:data-requirements}). Statistics are in Table~\ref{tab:corpus-stats}.
\begin{table*}[t]
\footnotesize
\centering
\resizebox{\textwidth}{!}{
\begin{tabular}{@{}lllrrrrrr@{}}
\toprule
\textbf{Corpus} & \textbf{Role} & \textbf{Domain} & \textbf{Docs} & \textbf{Segments} & \textbf{Src words} & \textbf{Systems} & \textbf{Term occ.} & \textbf{Glossary} \\
\midrule
\biomqm & validation (document) & biomedical & 50 & 384 & 7{,}536 & 10 & 1{,}320 & 729 \\
\STEP   & validation (occurrence) & geoscience & 10 & 4{,}432 & 124{,}285 & 1 & 3{,}377 & 17{,}035 \\
\ACL    & test & NLP (written) & 34 & 7{,}095 & 137{,}123 & 8 & 11{,}122 & 1{,}722 \\
\iwslt  & test & NLP (spoken) & 10 & 884 & 15{,}360 & 8 & 1{,}356 & 1{,}722 \\
\bottomrule
\end{tabular}}
\caption{Corpus statistics. Term occurrences are aligned term pairs per system for \biomqm, \ACL\ and \iwslt, and human-annotated occurrences for \STEP. All corpora are English--French.}
\label{tab:corpus-stats}
\end{table*}

\paragraph{\STEP\ (occurrence-level validation).}

\STEP\ \citep{carcamo:hal-05560658} contains 10~geoscience research articles translated from English into French by a document-level MT system fine-tuned on scientific abstracts of that domain \citep{peng-etal-2025-investigating}, and exhaustively annotated for terminology.
The associated resource is a 17{,}035-entry SKOS terminology compiled for the project from the Loterre thesauri\footnote{\url{https://www.loterre.fr}}  and the ARTES database \citep{pecman:hal-04996243}. Each occurrence of a known concept was extracted, aligned with its translation, and labelled with the fine-grained error typology of \citet{carcamo:hal-05560658}, the same typology our error judge instantiates (\S\ref{ssec:q1}), resulting in 3{,}377 gold-labelled occurrences (Appendix~\ref{app:dataset-details}). 27.7\%
of occurrences are found as is in the glossary, and 55.4\% are source-side variants of their concept's head term.
After projecting multi-label annotations onto the metric's label space, 69.8\% of occurrences are error-free, 11.0\% carry a family-A error, 16.2\% another error (C--H), and 3.0\% a sole in-document inconsistency label (B).

\paragraph{\biomqm\ (document-level validation).}

\biomqm\ \citep{zouhar-etal-2024-fine} is a multilingual biomedical benchmark annotated at the segment level with MQM error labels by professional translators. We use the English--French portion, 50 documents translated by 8~MT systems plus two human references. All MQM categories are retained when computing the human ground truth, because terminology-related problems are annotated under many categories, not only the Terminology label (below). Errors of other kinds, however, only count when they overlap with an aligned term occurrence (\S\ref{ssec:evaluation-methods}). The biomedical domain moreover exhibits a markedly higher rate of terminology errors than general news \citep{zouhar-etal-2024-fine}. 

MQM is a general error typology: \biomqm\ includes no term inventory, no occurrences and no source--target term alignments. Turning it into a terminology evaluation benchmark, \biomqmterms, is a contribution of this work. We manually extracted the biomedical terms of the 50 source documents by exhaustive reading, obtaining 729 concepts (26\% with at least one variant); French labels come from the bilingual MeSH thesaurus \citep{lipscomb-2000-mesh} where possible and from manual translation otherwise (construction details in Appendix~\ref{app:dataset-details}). The resulting glossary feeds
the pipeline of \S\ref{ssec:data-requirements}: 1{,}320 aligned occurrences per system, i.e.\ 13{,}200 term--translation pairs. 21.7\% of them are touched by at least one MQM error, of which only a fifth carry the Terminology label, confirming that terminology-related problems spill over into Accuracy, Linguistic and Style categories. \biomqmterms\ comprises the SKOS glossary and, for each system, the aligned pairs with their variation categories and MQM correspondences.

\paragraph{\ACL\ and \iwslt\ (test).}
We re-use the two corpora and MT outputs released by \citet{dahan-etal-2026-improving}, to which we refer for details, and re-run the preprocessing of \S\ref{ssec:data-requirements} on them. \ACL\ \citep{peng-etal-2026-parallel} is a corpus of NLP research papers published in *ACL venues and their comparable French versions.
\iwslt\ \citep{salesky-etal-2023-evaluating} comprises manually revised transcripts and translations of 10 presentations delivered at ACL 2022, from the IWSLT 2023 shared task. Both corpora share the INIST/Loterre NLP glossary of 1{,}722 EN--FR concepts.\footnote{\url{https://skosmos.loterre.fr/8LP/fr/}} 
MT outputs are generated by four open-weight LLMs, each producing two translations of every document: one from plain translation instructions (\emph{baseline}) and one with the glossary-preferred terms in the segment and their French translations injected into the prompt (\emph{base+terms}), yielding eight outputs per corpus.

\subsection{Occurrence-Level Validation on \STEP}
\label{ssec:step-protocol}

For \STEP, \method\ is run on the gold-aligned occurrences, and its verdicts are compared with the human labels occurrence by occurrence. Since the human annotation uses the full 23-code typology while the judge uses its simplified version (\S\ref{ssec:q1}), the gold codes are grouped in the same way, so that both sides share 15 classes: \Aone--\Aseven, \textbf{C}--\textbf{H} (Table~\ref{tab:verdicts}), \textbf{B} (unjustified in-document variation) and no-error.
As human annotations are multi-label (25.6\% of occurrences carry at least two codes), they are reduced to one label with a fixed precedence (the most severe A code, then the C--H code, then B, pure inconsistency, then no-error). Consistency notes on locally correct translations (B1/B2) reduce to no-error at the occurrence level, and 4~occurrences with missing annotations are excluded, leaving 3{,}373 scored occurrences. 
We report per-class precision, recall and F1, macro-F1, accuracy and balanced accuracy, together with a binary error-detection view in which all error types collapse to a single error class. On the metric side, \VarNJ\ maps to B (both denote in-document inconsistency) while \Conforme\ and \VarJ\ map to no-error, the conformity/variation distinction being absent from the human annotation.

\subsection{Document-Level Meta-Evaluation with \biomqmterms}
\label{ssec:evaluation-methods}

\noindent\textbf{Human ground truth per (document, system).}
Human terminology quality is derived from the MQM annotations at the level of aligned term pairs, not at the level of full segments. For each alignment produced by our pipeline, we aggregate the severity weights of all MQM errors overlapping the target span, using the standard WMT weighting (Critical: 25, Major: 5, Minor: 1, Neutral: 0). When a segment is seen by multiple annotators, the penalty is averaged at the annotator level (annotators who saw the segment but flagged no error contribute 0). The document-level human score is the negated mean alignment penalty (higher is better).

\noindent\textbf{Meta-evaluation measures.}
We evaluate the metrics against human scores at three levels using the \texttt{mt-metrics-eval} package.\footnote{\url{https://github.com/google-research/mt-metrics-eval}} At the system level, Soft Pairwise Accuracy (SPA) \citep{thompson-etal-2024-improving} checks, for each of the 45 system pairs, whether the metric separates the two systems with the same confidence as the human scores (p-values of paired permutation tests over documents, 1{,}000 permutations), and averages this agreement over the pairs. At the segment level, the group-by-item pairwise accuracy with tie calibration of \citet{deutsch-etal-2023-ties}, \textbf{\acceq}, treats each of the 373 source segments carrying at least one aligned pair (out of the corpus's 384) as an item (a 10-system $\times$ 373-segment score matrix); ties matter at this level, as for 29\% of the segments all ten systems receive the same human score. At the document level, the same measure over the 50~documents (a $10 \times 50$ matrix) is reported for completeness: human document scores, averaged over about 26 occurrences, are practically never tied, so the measure reduces to plain pairwise accuracy over 50 items.

\noindent\textbf{Score aggregation and comparison scope.}
Because \method\ assigns a judgement to every occurrence (\S\ref{ssec:pipeline}), its document score can be aggregated in three ways: \emph{micro-averaged} over occurrences, and \emph{macro-averaged} per lemmatised source form or per concept; we report all three. Comparison metrics score what they are designed for: the glossary-conformity and divergence-based baselines score the same aligned occurrences as \method, while the terminology-filtered and general QE baselines score every segment of the document. In a \emph{no-hapax} variant, only chains with at least two occurrences are evaluated, for both the metrics and the human scores. This leaves 48 documents; the segment-level QE baselines, which do not depend on term occurrences, keep the full 50 documents. The results of this no-hapax variant are given in Appendix~\ref{app:nohapax} (Tables~\ref{tab:nohapax-occ}--\ref{tab:nohapax-concept}).

\noindent\textbf{Comparing systems.}
\method\ is compared against four metric families, ordered from most to least terminology-aware. (i)~\emph{Divergence QE}: the same pipeline through divergence detection, with typed judges replaced by a quality-estimation model scoring each divergent term pair (the source term and its aligned translation, without the surrounding sentence) (CometKiwi \citep{rei-etal-2022-cometkiwi}, MetricX-24 \citep{juraska-etal-2024-metricx} and \textsc{Gemba-MQM} \citep{kocmi-federmann-2023-gemba}; subscripted \emph{div}), which isolates the judges’ contribution from that of the divergence architecture. (ii)~\emph{Glossary conformity} 
in the consistency tradition \citep{alam2021evaluationmachinetranslationterminology,semenov-bojar-2022-automated}:
the head term is matched against its glossary label, and the remaining forms, those without an official reference, against a document-level \emph{pseudo-reference} (first or majority translation), on lemmas. 
(iii)~\emph{Terminology-filtered QE}, a naive LLM judge: \textsc{Gemba-MQM} restricted to the errors flagged as terminological (by category, or by any mention of terminology). (iv)~\emph{General QE}: the same backbones scoring every sentence in the document with no notion of term (subscripted \emph{gen}), averaged over the document, as a generic baseline.

\subsection{System Ranking on the Test Corpora}
\label{ssec:test-protocol}

On \ACL\ and \iwslt, \method\ runs in the configuration of \S\ref{ssec:default-config} with the domain set to NLP, and scores the eight MT outputs distributed with the corpora: four open-weight LLMs (Llama-3.1-8B-Instruct \citep{grattafiori-etal-2024-llama3}, Qwen3-8B \citep{yang-etal-2025-qwen3}, EuroLLM-9B-Instruct \citep{martins-etal-2025-eurollm9b} and EuroLLM-22B-Instruct \citep{ramos-2026-etal-eurollm22b}), each under the two zero-shot conditions of \S\ref{sec:datasets}, \emph{baseline} and \emph{base+terms}. Decoding is greedy except for Qwen3-8B, run in its recommended non-thinking sampling mode. 

\section{Results and Analyses}
\label{sec:results}

\subsection{Occurrence-Level Agreement on \STEP}
\label{ssec:results-step}

Table~\ref{tab:step-results} reports agreement with the gold labels for 3{,}373 \STEP\ occurrences under the \emph{exhaustive judging path} (Appendix~\ref{app:exhaustive}). The deterministic matching shortcut is disabled: every occurrence, including matches and hapaxes, is submitted to Q1 and, if error-free, Q2, allowing every gold class to be predicted.

% \begin{table}[!tbp]
% \footnotesize
% \centering
% \setlength{\tabcolsep}{4pt}
% \begin{tabular}{@{}lc@{}}
% \toprule
%  & \method \\
% \midrule
% \multicolumn{2}{@{}l@{}}{\emph{15-class agreement}} \\
% Macro-F1            & 0.196 \\
% Accuracy            & 0.644 \\
% Balanced accuracy   & 0.230 \\
% A-error detection   & 0.590 \\
% Exact A-code (among detected) & 0.521 \\
% \midrule
% \multicolumn{2}{@{}l@{}}{\emph{Binary error detection}} \\
% F1 (ERROR)          & 0.620 \\
% Precision (ERROR)   & 0.632 \\
% Recall (ERROR)      & 0.609 \\
% Accuracy            & 0.775 \\
% Balanced accuracy   & 0.728 \\
% \midrule

% \multicolumn{2}{@{}l@{}}{\emph{Binary accuracy by hapax status}} \\
% Hapax ($n{=}339$)        & 0.664 \\
% Non-hapax ($n{=}3{,}034$) & 0.787 \\
% \bottomrule
% \end{tabular}
% \caption{Agreement between \method's verdicts and the human gold labels on \STEP, every occurrence being judged through the exhaustive path (\S\ref{ssec:results-step}). 3{,}373 scored occurrences: 1{,}019 ERROR and 2{,}354 NO-ERROR, of which 371 carry a family-A label.}
% \label{tab:step-results}
% \end{table}

\begin{table}[!tbp]
\footnotesize
\centering
\setlength{\tabcolsep}{4pt}
\begin{tabular}{@{}lc@{}}
\toprule
 & \method \\
\midrule
\multicolumn{2}{@{}l@{}}{\emph{15-class agreement}} \\
Macro-F1                      & 0.196 \\
Accuracy / Balanced accuracy  & 0.644 / 0.230 \\
A-error detection             & 0.590 \\
Exact A-code (among detected) & 0.521 \\
\midrule
\multicolumn{2}{@{}l@{}}{\emph{Binary error detection}} \\
F1 / Precision / Recall (error class) & 0.620 / 0.632 / 0.609 \\
Accuracy / Balanced accuracy   & 0.775 / 0.728 \\
\midrule
\multicolumn{2}{@{}l@{}}{\emph{Binary accuracy by hapax status}} \\
Hapax ($n{=}339$)          & 0.664 \\
Non-hapax ($n{=}3{,}034$) & 0.787 \\
\bottomrule
\end{tabular}
\caption{Agreement between \method\ and human gold labels on \STEP, all occurrences judged via the exhaustive path (\S\ref{ssec:results-step}): 3{,}373 scored occurrences, including 1{,}019 errors (371 with a family-A label) and 2{,}354 non-errors.}
\label{tab:step-results}
\end{table}

\method\ detects errors well above the majority-class baseline (balanced accuracy 0.728). Fine-grained code assignment remains hard: 59\% of the gold family-A occurrences are flagged with some A code but only about half receive the correct code, with a 15-class macro-F1 close to 0.20. 
For hapax occurrences (339), binary accuracy drops by about 12~points relative to non-hapax occurrences (0.664 vs.\ 0.787, Table~\ref{tab:step-results}), quantifying the difficulty of judging terms without document-internal evidence. 
Three patterns dominate the confusion matrix (Appendix~\ref{app:confusion}, Table~\ref{tab:confusion}). First, the missed errors are mostly not terminological errors proper: of the 398~gold errors judged correct, 159 belong to family F (grammar around the term: agreement, determiners, prepositions) and 82 to pure in-document inconsistency (B), two families that concern the term's surroundings or its consistency rather than the choice of equivalent. Among wrong-equivalent errors (family A), the weak point is A3, the invented calque, accepted as correct in 40 of 65 cases (one is shown in Table~\ref{tab:qual-examples}): the judge takes a literal translation for an attested term. Second, content-altering translations (G: omissions, additions, unintelligible output) are almost always caught (215 out of 227 are tagged erroneous) but mostly with a wrong label (126 as A1): the judge detects that a translation is wrong more reliably than it identifies the mechanism, and this family-level confusion explains the gap between binary balanced accuracy (0.728) and low 15-class macro-F1 (0.196). Third, the gold and judged notions of unjustified variation barely overlap: only 4 of the 101 pure-inconsistency cases are flagged \VarNJ, while 104 of the 149 \VarNJ\ verdicts were not tagged by the annotators. This reflects a difference of status. In the human guidelines, variation between individually correct translations is recorded (family B) but not counted as an error and carries no severity (Appendix~\ref{app:typology}); \method, by contrast, penalises such variation when the consistency judge finds it unjustified (\VarNJ, Table~\ref{tab:verdicts}). The human B labels thus describe variation without judging whether it is justified, whereas \VarNJ\ asserts that it is not, so the two capture different cases. The LLM judge and the human annotators therefore disagree most on the consistency axis, not on error detection.

\subsection{Document-Level Meta-Evaluation}
\label{ssec:results-biomqm}

Table~\ref{tab:biomqm-occ} reports the meta-evaluation on \biomqmterms\ under the micro-average over occurrences defined in \S\ref{ssec:aggregation}. Macro-averaging per source form or concept yields the same conclusions (Appendix~\ref{app:macro-agg}).
\method\ ranks first in SPA (0.844) and in segment-level \acceq\ (0.606). On segment-level \acceq, its lead is significant over every other metric ($p \leq 0.01$, paired permutation tests). On SPA, it is significant for eight of the ten metrics ($p < 0.05$, $k{=}2{,}000$), with ties for the Divergence-QE variants \textsc{Gemba}$_\text{div}$ and CometKiwi$_\text{div}$. 
The Divergence-QE variants form the second-best family; the remaining families follow within a narrow band (SPA 0.69--0.78).

The document-level \acceq\ column, where the measure reduces to plain pairwise accuracy over 50 items (\S\ref{ssec:evaluation-methods}), is reported for completeness only. Excluding hapax chains removes occurrences lacking another rendering in the document; unless the term is in the glossary, only a judgement of the translation itself, by an LLM or a human, can evaluate them. In that reduced variant, no SPA difference between \method\ and the remaining metrics is statistically significant ($p \geq 0.17$), while \method\ stays significantly ahead of general QE on segment-level \acceq\ (Appendix~\ref{app:nohapax}).

\begin{table}[!tbp]
\footnotesize
\centering
\setlength{\tabcolsep}{3.5pt}
\resizebox{\columnwidth}{!}{%
\begin{tabular}{@{}lccc@{}}
\toprule
\textbf{Metric} & \textbf{SPA} & \textbf{\acceq} (seg) & \textbf{\acceq} (doc) \\
\midrule
\multicolumn{4}{@{}l}{\emph{Judged divergences (this work)}} \\
\textbf{\method}            & \textbf{0.844} & \textbf{0.606} & \textbf{0.557} \\
\multicolumn{4}{@{}l}{\emph{Divergence QE}} \\
\quad \textsc{Gemba}$_\text{div}$        & 0.803$^{\phantom{*}}$ & 0.597$^{*}$ & 0.516 \\
\quad MetricX-24$_\text{div}$            & 0.769$^{*}$ & 0.593$^{*}$ & 0.526 \\
\quad CometKiwi$_\text{div}$             & 0.786$^{\phantom{*}}$ & 0.599$^{*}$ & 0.543 \\
\multicolumn{4}{@{}l}{\emph{Glossary conformity}} \\
\quad first-translation fallback & 0.730$^{*}$ & 0.595$^{*}$ & 0.449 \\
\quad majority fallback     & 0.723$^{*}$ & 0.598$^{*}$ & 0.445 \\
\multicolumn{4}{@{}l}{\emph{Terminology-filtered QE}} \\
\quad \textsc{Gemba-MQM} (any mention) & 0.775$^{*}$ & 0.586$^{*}$ & 0.219 \\
\quad \textsc{Gemba-MQM} (category)    & 0.689$^{*}$ & 0.583$^{*}$ & 0.193 \\
\multicolumn{4}{@{}l}{\emph{General QE}} \\
\quad \textsc{Gemba}$_\text{gen}$        & 0.736$^{*}$ & 0.583$^{*}$ & 0.533 \\
\quad MetricX-24$_\text{gen}$            & 0.736$^{*}$ & 0.586$^{*}$ & 0.566 \\
\quad CometKiwi$_\text{gen}$             & 0.729$^{*}$ & 0.591$^{*}$ & 0.552 \\
\bottomrule
\end{tabular}%
}
\caption{Meta-evaluation on \biomqmterms, \textbf{micro-average} occurrence aggregation (10 systems, 50 documents, 373 segments; human ground truth from all MQM categories). Empirical SPA chance level: $0.592 \pm 0.089$. $^{*}$: \method\ is significantly better (paired permutation tests, $p<0.05$); unmarked values are ties.}
\label{tab:biomqm-occ}
\end{table}

Like any typology-based metric, the document score relies on a penalty scale, which follows the gravities assigned by the typology's authors, each grouped family C--H being scored at its most lenient sub-code (benefit of the doubt, Table~\ref{tab:verdicts}). Varying C, D and E weights from 1 to 4 with $F{=}G{=}H{=}1$ leaves \method\ ahead of all comparison metrics under occurrence and source-form aggregation (Appendix~\ref{app:penalties}).

\begin{table*}[t]
\footnotesize
\centering
\setlength{\tabcolsep}{4pt}
\resizebox{\textwidth}{!}{%
\begin{tabular}{@{}lllll@{}}
\toprule
\textbf{Source form} & \textbf{MT output} & \textbf{Judgement} & \textbf{Verdict} & \textbf{Judge's reason (abridged)} \\
\midrule
\emph{annotation}$^\star$ & annotation humaine & explicitation & \VarJ & first mention, expands the term with a clarifying adjective \\
\emph{dataset} & jeu de données & synonym inconsistency & \VarNJ & document established \emph{ensemble de données}; switch without visible reason \\
\emph{attention} & auto-attention & neutralisation & \VarNJ & reproduces the translation reserved for another source form of the concept \\
\emph{word embedding} & incorporation de mots & invented calque & \Athree & expected \emph{plongement lexical}; calque not attested in the domain \\
\emph{retrieval} & récupération & general word & \Aseven & expected \emph{recherche}; general word, not the specialised term \\
\midrule

\emph{rate and state friction}$^\star$ & friction à taux et état & first mention & \VarJ\ (gold: \Athree) & the calque is an established French term, attested in the domain\\
\bottomrule
\end{tabular}%
}
\caption{Five judgements on one system output (EuroLLM-22B-Instruct, \emph{baseline}, \iwslt) and, below the rule, a documented judgement error from \STEP\ (human gold: \Athree). The last column is the explanation the judge returns with each verdict, abridged. $^\star$: first mention.}
\label{tab:qual-examples}
\end{table*}

\subsection{System Ranking on \ACL\ and \iwslt}
\label{ssec:results-tal}

Table~\ref{tab:tal-results} ranks the eight systems of \S\ref{sec:datasets} on both corpora with \method\ and with two of the comparison families of \S\ref{ssec:evaluation-methods}. With \method, for all four models the \emph{base+terms} run scores better than its \emph{baseline} run, on both corpora, with the largest gains on \iwslt. Where \citet{dahan-etal-2026-improving} observed a tension they could not arbitrate (glossary injection improves accuracy and consistency counts while reducing the transfer of source variation), \method\ returns a clearer verdict, and its verdict distribution (Table~\ref{tab:tal-profiles}) shows why: under \emph{base+terms}, exact conformity rises by 6 points on \ACL\ and 12 on \iwslt, wrong-equivalent (A) error rates drop sharply (11.7\% to 8.4\% of occurrences on \ACL, 18.6\% to 9.3\% on \iwslt), while justified and unjustified variation both recede moderately; the variation that glossary injection suppresses is mostly not valid, so \emph{base+terms} yields a net gain on these corpora.
The two comparison metrics behave on these corpora as they did in the meta-evaluation. General QE is blind to the terminological gain: CometKiwi$_\text{gen}$ orders the systems by general quality, is nearly insensitive to the prompting condition, prefers the \emph{baseline} run in most pairs of one \emph{baseline} and one \emph{base+terms} run (eleven of sixteen on \iwslt, ten on \ACL\ with one tie), and its best run for each corpus is a \emph{baseline} run. Glossary conformity ranks every \emph{base+terms} run above every \emph{baseline} run (all sixteen pairs on each corpus), as it counts exactly the glossary terms that the \emph{base+terms} prompt injects, but disagrees with \method\ within each condition: on \iwslt\ it ranks Qwen3-8B first among the \emph{baseline} runs, where \method\ ranks it last. Conformity only checks whether the glossary term appears in the output; judges also recognise justified variation and detect errors that matching approaches cannot.

\begin{table}[t]
\footnotesize
\centering
\setlength{\tabcolsep}{2pt}
\resizebox{\columnwidth}{!}{%
\begin{tabular}{@{}lcccccccc@{}}
\toprule
& \multicolumn{4}{c}{\emph{baseline}} & \multicolumn{4}{c}{\emph{base+terms}} \\
\cmidrule(lr){2-5}\cmidrule(lr){6-9}
\textbf{Metric} & Ll. & Qw. & E9 & E22 & Ll. & Qw. & E9 & E22 \\
\midrule
\multicolumn{9}{@{}l}{\ACL} \\
\midrule
\method\,$\downarrow$ & .322 & .350 & .287 & .312 & .220 & \textbf{.214} & .236 & .224 \\
Conformity\,$\uparrow$ & .773 & .768 & .776 & .783 & .851 & \textbf{.864} & .826 & .844 \\
CometKiwi$_\text{gen}$\,$\uparrow$ & .828 & .834 & \textbf{.839} & .838 & .823 & .829 & .838 & .835 \\
\midrule
\multicolumn{9}{@{}l}{\iwslt} \\
\midrule
\method\,$\downarrow$ & .468 & .502 & .437 & .439 & .236 & .245 & .330 & \textbf{.230} \\
Conformity\,$\uparrow$ & .713 & .723 & .719 & .731 & .848 & \textbf{.850} & .788 & .833 \\
CometKiwi$_\text{gen}$\,$\uparrow$ & .807 & .815 & \textbf{.823} & \textbf{.823} & .803 & .808 & .821 & .818 \\
\bottomrule
\end{tabular}%
}
\caption{System comparison on the two NLP test corpora with \method\ (mean per-document penalty, lower is better), glossary conformity and CometKiwi general QE (families~(ii) and~(iv) of \S\ref{ssec:evaluation-methods}); best value per metric and corpus in bold. Ll.: Llama, Qw.: Qwen, E9/E22: EuroLLM-9B/22B.}
\label{tab:tal-results}
\end{table}

\begin{table}[t]
\footnotesize
\centering
\setlength{\tabcolsep}{4pt}
\resizebox{\columnwidth}{!}{%
\begin{tabular}{@{}llccccc@{}}
\toprule
\textbf{Corpus} & \textbf{Condition} & \textbf{Conf.} & \textbf{Just.} & \textbf{Unjust.} & \textbf{A} & \textbf{C--H} \\
\midrule
\ACL   & baseline   & 72.2 & 11.6 & 1.6 & 11.7 & 2.9 \\
       & base+terms & 78.5 &  9.2 & 1.3 &  8.4 & 2.6 \\
\midrule
\iwslt & baseline   & 67.5 &  9.2 & 2.2 & 18.6 & 2.5 \\
       & base+terms & 79.2 &  7.5 & 2.1 &  9.3 & 1.9 \\
\bottomrule
\end{tabular}%
}
\caption{Distribution of verdicts by prompting condition (percentage of occurrences, four models pooled): conformity, justified and unjustified variation, wrong-equivalent errors (A) and other error families (C--H).}
\label{tab:tal-profiles}

\end{table}

\subsection{What Judging Adds over Counting}

\label{sec:analysis}

\paragraph{The comparison metrics as an ablation of \method.}
Read from the bottom of Table~\ref{tab:biomqm-occ} upwards, the comparison families of \S\ref{ssec:evaluation-methods} add one component of \method\ at a time, so the table can be viewed as an ablation study. General QE is the baseline (SPA 0.729--0.736), restricting it to terminology-mentioning error spans changes little and can even hurt (0.775 for the permissive filter, 0.689 for the strict one), and counting glossary conformity instead of judging results in similar scores (0.723--0.730). The first substantial step is architectural: keeping the document-level divergence detection but scoring the divergence points with an off-the-shelf QE metric increases the score to 0.769--0.803; the divergence architecture already carries part of the signal. The second step is the judgement itself: replacing the generic score at those points with the typed Q1/Q2 judges adds the rest (0.844), and it is the only step that also provides interpretable labels.

\paragraph{What consistency labels reveal.}

Beyond the binary \VarJ/\VarNJ\ outcome, the labels returned by the Q2 judge also say why a divergence is judged coherent or not (Appendix~\ref{app:q2-labels}, Table~\ref{tab:q2-labels}). Their distribution is stable across systems: about four divergences out of five follow the document's established usage (e.g., \emph{treatment} rendered \emph{traitement} throughout while the glossary reference is \emph{thérapeutique}: divergent at every occurrence, yet error-free and tagged as `document usage', \VarJ); `first mention' accounts for another 6--7\%.

The two prompting conditions differ where it matters. Under \emph{base+terms}, `synonym inconsistency' goes down for every model (3.0--3.7\% to 2.0--2.9\%) while `neutralisation' goes up slightly (0.8--1.8\% to 1.5--2.1\%): glossary injection eliminates unmotivated switching between synonyms, but slightly more often erases a source distinction between a term and one of its variants. This is the loss of source-side variation already observed in \S\ref{ssec:results-tal}, and it remains far smaller than the reduction of genuine errors (Table~\ref{tab:tal-profiles}; the same pattern holds on \iwslt, where small counts call for caution).

\paragraph{Example Judgements.}

Table~\ref{tab:qual-examples} displays five real judgements from a single system output. The same situation, a translation that differs from the document's dominant one, is judged in four different ways depending on whether it erases a source distinction, switches designation without purpose, expands the term at its first mention, or is simply a wrong term. The judge also returns an explanation, which makes each decision interpretable.
The last row, from \STEP, is a judge error of the kind quantified in \S\ref{ssec:results-step} (40 of 65 invented calques accepted): the judge does not merely accept the invented calque, it asserts that it is attested.

\section{Conclusion}
\label{sec:conclusion}

\method\ starts from a simple observation: terminology metrics that count divergences from a fixed reference conflate translation errors with valid variations that translators routinely produce. It therefore judges instead of counting: matches with the expected translation are settled deterministically, divergences go, in document context, to an error judge and then to a consistency judge, and the interpretable verdicts aggregate into a document-level score. On \STEP, error detection agrees with a per-occurrence expert gold (binary balanced accuracy 0.728); on \biomqmterms, \method\ ranks first in SPA under all three aggregations and in segment-level \acceq, significantly ahead of every comparison metric at that level. On two NLP test corpora, it settles a question counting could not: glossary injection helps every model because it removes genuine errors, not legitimate variation. The pipeline carries over to other domains given a glossary and aligned occurrences, and to other language pairs by adapting its prompts; testing open-weight judges, and the stability of the verdicts across judge models and prompts, are natural next steps.

\section*{Limitations}
\label{sec:limitations}

\method\ inherits the limitations of LLM-as-a-judge evaluation. Verdicts depend on a single proprietary judge model (\texttt{gpt-4.1-mini}, chosen as explained in \S\ref{ssec:default-config}); all prompts are versioned, but the stability of the verdicts across judge models and prompt paraphrases was not measured and no open-weight judge was tested, although the conclusions hold across judging paths, penalty scales and aggregations (Appendices~\ref{app:exhaustive}, \ref{app:penalties} and~\ref{app:macro-agg}), and exposing the full 23-code typology instead of the compact one moves binary error detection by under three points (Appendix~\ref{app:typology-comparison}). Pseudo-references are derived from the very output under evaluation; the validation by the error judge and the LLM generation of Step~1 mitigate this circularity but do not eliminate it. Out-of-glossary hapaxes remain the hardest case, with binary agreement about 12 points below non-hapax occurrences on \STEP. Ignoring diacritics and hyphens in Step~2 can also conflate distinct words offered for the same term (\emph{élevé} vs \emph{élève}).

The validation supports the metric at the granularities at which it is used, binary error detection and aggregated document scores; per-code agreement is substantially lower (15-class macro-F1 0.196), mostly because the judge detects that a translation is wrong more reliably than it identifies the mechanism (\S\ref{ssec:results-step}). Fine-grained codes are therefore reported for diagnosis and should be read as indicative; the document score, for its part, is robust to the weights assigned to the grouped families (Appendix~\ref{app:penalties}). Occurrence-level validation relies on consensus labels from two annotators working together and on a single MT system on \STEP, so annotator bias cannot be separated from metric error, and document-level validation is limited to one language pair (EN--FR) and ten systems on \biomqm; the judges' accuracy may differ for other language pairs and domains.

Finally, \method\ requires a domain glossary and an upstream pipeline (head term and variant detection, variation labelling, alignment) whose errors propagate to the verdicts. This propagation was not measured directly, but the two validations bracket it: \STEP\ evaluates the judges alone on gold occurrences and alignments, whereas \biomqmterms\ is processed end-to-end by the pipeline, and \method\ still ranks first there. The upstream steps were checked on samples: the variation labelling in prior work, with an earlier prompt, and term detection and the LLM alignment on \ACL\ only, where excluding the least reliable detections and alignments leaves the rankings unchanged (Appendices~\ref{app:pipeline} and~\ref{app:preproc-validation}). Document scores are also relative to the terminology provided, so comparisons are meaningful across systems evaluated on the same corpus and terminology, not across corpora. Evaluating a full corpus incurs a non-trivial LLM cost (Appendix~\ref{app:cost}).

\section*{Acknowledgments}

This work was supported by the French National Agency (ANR) as part of the MaTOS project under reference ANR-22-CE23-0033.\footnote{\url{http://anr-matos.github.io/}} The authors are also grateful to the anonymous reviewers for their insightful comments and suggestions, and to Natalie Kübler, Alexandra Mestivier and José Cornejo C\'arcamo for early discussions on term variation and evaluation of term translation, to \'Eric de la Clergerie for his help on the \Concordancer and to the MaTOS project team for their feedback on preliminary versions of this work.

\bibliography{custom}

\appendix

\section{Pipeline Overview and a Worked Example}
\label{app:pipeline}

\begin{figure*}[t]
\centering
\definecolor{ppInk}{HTML}{263238}%
\definecolor{ppGrey}{HTML}{78909C}%
\definecolor{ppRule}{HTML}{CFD8DC}%
\definecolor{ppPre}{HTML}{3E5C76}%
\definecolor{ppPreL}{HTML}{E9EFF5}%
\definecolor{ppMet}{HTML}{00796B}%
\definecolor{ppMetL}{HTML}{E0F2F1}%
\definecolor{ppIn}{HTML}{F4F6F7}%
\definecolor{ppInB}{HTML}{B0BEC5}%
\definecolor{ppOk}{HTML}{2E7D32}%
\definecolor{ppOkBg}{HTML}{E8F5E9}%
\definecolor{ppBad}{HTML}{C62828}%
\definecolor{ppBadBg}{HTML}{FFEBEE}%
\def\llmbadge{{\setlength{\fboxsep}{1.2pt}\colorbox{ppMet}{\color{white}\fontsize{4.8}{5.5}\selectfont\bfseries LLM}}}%
\begin{tikzpicture}[
    font=\fontsize{6}{6.9}\selectfont, every node/.style={align=center},
    inbox/.style={draw=ppInB, fill=ppIn, rounded corners=2pt, line width=0.5pt, text=ppInk,
                  inner xsep=2pt, inner ysep=2.5pt, text width=0.95cm, font=\fontsize{6}{6.9}\selectfont\itshape},
    box/.style={rounded corners=2pt, line width=0.5pt, text=ppInk,
                minimum width=1.35cm, minimum height=0.9cm, inner xsep=2pt, inner ysep=2pt},
    pre/.style={box, draw=ppPre, fill=white},
    met/.style={box, draw=ppMet, fill=white},
    okpill/.style={draw=ppOk, fill=ppOkBg, text=ppOk, rounded corners=4pt, line width=0.5pt,
                   font=\fontsize{5.8}{6.8}\selectfont\bfseries, inner xsep=2.5pt, inner ysep=2pt, anchor=west},
    badpill/.style={draw=ppBad, fill=ppBadBg, text=ppBad, rounded corners=4pt, line width=0.5pt,
                    font=\fontsize{5.8}{6.8}\selectfont\bfseries, inner xsep=2.5pt, inner ysep=2pt, anchor=west},
    flow/.style={-{Stealth[length=3pt, width=2.4pt]}, draw=ppGrey, line width=0.5pt, rounded corners=1.5pt},
    gline/.style={draw=ppGrey, line width=0.5pt, rounded corners=1.5pt},
    mflow/.style={-{Stealth[length=3pt, width=2.4pt]}, draw=ppMet, line width=0.5pt, rounded corners=1.5pt},
    hline/.style={draw=ppMet, line width=0.5pt, dash pattern=on 2pt off 1.3pt},
    mline/.style={draw=ppMet, line width=0.5pt},
    lbl/.style={font=\fontsize{6}{7}\selectfont\itshape, text=ppInk!75, inner sep=1pt},
    grp/.style={font=\fontsize{6.5}{7.5}\selectfont, inner sep=0pt, align=left}
]
% ---- budget horizontal (cm), boites uniformes de 1.35 x 0.90, tout sur une ligne :
%      entrees 0.02-1.11 | bus 1.25 | panneau pretraitement 1.40-6.40 : p1 1.55-2.92,
%      p2 3.24-4.61, p3 4.93-6.30 | panneau TermJudge 6.52-16.00 : ref 6.62-7.99,
%      div 8.27-9.64, Q1/Q2 10.56-11.93 | bifurcation 12.09 | verdicts des 12.56 |
%      collecteur 14.28 | score 14.55-15.92
\def\xv{12.555}   % bord gauche des verdicts
\def\xk{14.28}   % collecteur vertical
% panneaux des deux etages
\fill[ppPreL, rounded corners=4pt] (1.40,-2.10) rectangle (6.40, 1.95);
\node[grp, text=ppPre, anchor=north west] at (1.54, 1.88) {\textls[60]{\textsc{Preprocessing: document-level}}\\[-1pt]\textls[60]{\textsc{term annotation pipeline}}};
\fill[ppMetL, rounded corners=4pt] (6.52,-2.10) rectangle (16.00, 1.95);
\node[grp, text=ppMet, anchor=north west] at (6.66, 1.88) {\textls[70]{\method\ (this work)}};
% entrees, fusionnees sur un bus vertical avant la premiere etape
\node[inbox] (gloss) at (0.565, 0.60) {glossary};
\node[inbox] (docs)  at (0.565,-0.65) {source\\documents\\+ MT outputs};
\node[pre] (p1) at (2.235, 0) {term\\\& variant\\spotting};
\node[pre] (p2) at (3.925, 0) {variation\\labelling};
\node[pre] (p3) at (5.615, 0) {term\\alignment +\\normalisation};
\draw[gline] (docs.east)  -- (1.25,-0.65) -- (1.25, 0.60) -- (gloss.east);
\draw[flow]  (1.25, 0) -- (p1.west);
\draw[flow] (p1) -- (p2);
\draw[flow] (p2) -- (p3);
% etapes de TermJudge
\node[met] (ref)   at (7.30, 0)     {reference\\selection\\$R(s)$};
\node[met] (div)   at (8.95, 0)     {divergence\\detection};
\node[met] (q1)    at (11.24, 0)    {\llmbadge\\[1pt]error\\judge Q1};
\node[met] (q2)    at (11.24,-1.25) {\llmbadge\\[1pt]consistency\\judge Q2};
\node[met] (score) at (15.235, 0)    {penalties $\to$\\document\\score};
\draw[mflow] (p3) -- (ref);
\node[lbl, anchor=north east] at (p3.south east |- 0,-0.55) {aligned occurrences $(s \to t)$};
\draw[mflow] (ref) -- (div);
% verdicts
\node[okpill]  (conf)  at (\xv, 1.25) {\Conforme};
\node[badpill] (err)   at (\xv, 0.00) {A1..A7, C..H};
\node[okpill]  (varj)  at (\xv,-0.90) {\VarJ};
\node[badpill] (varnj) at (\xv,-1.60) {\VarNJ};
% arbre de decision
\draw[mflow] (div.north) |- node[lbl, above, pos=0.78] {match} (conf.west);
\draw[mflow] (div.east) -- node[lbl, above] {divergence\\or hapax} (q1.west);
\draw[mflow] (q1.east) -- node[lbl, above] {error} (err.west);
\draw[mflow] (q1.south) -- node[lbl, left] {no error} (q2.north);
\draw[hline] (q1.north) -- node[lbl, right] {no error\\(hapax)} (q1.north |- 0,1.25);
\fill[ppMet] (q1.north |- 0,1.25) circle (1.2pt);
\draw[mflow] (q2.east) -- ++(0.16,0) |- (varj.west);
\draw[mflow] (q2.east) -- ++(0.16,0) |- (varnj.west);
\node[lbl, anchor=south] at (\xv-0.15,-0.90+0.03) {yes};
\node[lbl, anchor=north] at (\xv-0.15,-1.60-0.03) {no};
% collecteur des verdicts vers le score
\draw[mline] (\xk, 1.25) -- (\xk,-1.60);
\foreach \n in {conf, err, varj, varnj} { \draw[mline] (\n.east) -- (\n.east -| \xk,0); \fill[ppMet] (\n.east -| \xk,0) circle (1.2pt); }
\draw[mflow] (\xk, 0) -- (score.west);
\end{tikzpicture}
\caption{Overview of the evaluation pipeline (every data-side notion is defined in \S\ref{ssec:data-requirements}). Upstream: the \Concordancer\ spots the occurrences of glossary concepts and of their variants in the source documents; each source form is labelled with a variation category; each occurrence is aligned with its translation, normalised into a canonical form and a lemma. \method\ (this work): a reference $R(s)$ is fixed per source form and document; a translation matching $R(s)$ is \Conforme\ without any LLM call; a divergent translation is classified by the LLM error judge (Q1), which either assigns an error label (\Aone--\Aseven, \textbf{C}--\textbf{H}) or passes it to the LLM consistency judge (Q2), which returns \VarJ\ or \VarNJ; hapax occurrences are always submitted to Q1 and, if error-free, labelled \Conforme\ without the consistency stage; individual penalties aggregate into a document-level score.}
\label{fig:pipeline}
\end{figure*}
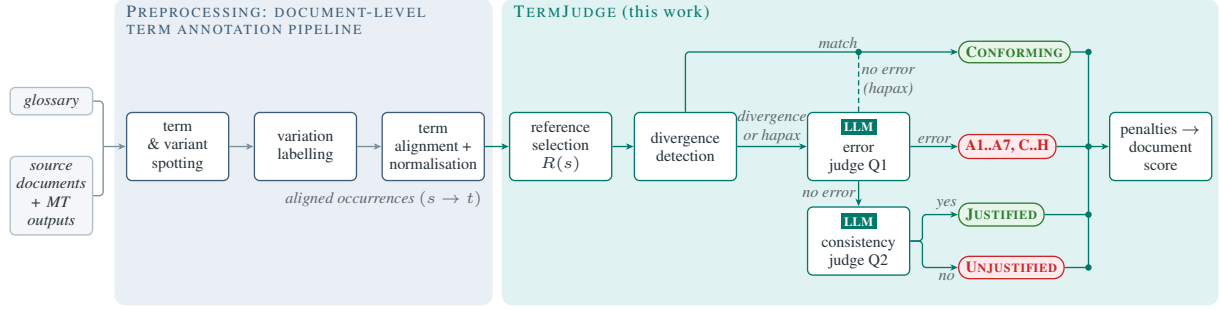

Figure~\ref{fig:pipeline} situates \method\ within the full evaluation pipeline: the upstream stages produce the aligned term occurrences defined in \S\ref{ssec:data-requirements}, and the metric proper begins at reference selection. These stages are the preprocessing pipeline of \citet{dahan-etal-2026-improving}, which we run ourselves on \biomqmterms, \ACL\ and \iwslt; \STEP\ comes with gold occurrences and alignments. In the three corpora, the MT outputs follow the source segmentation, so that source and target segments are aligned by construction. Variation labelling uses a revised version of that study's prompt (Appendix~\ref{app:prompt-variation}), run with \texttt{gpt-4.1}. The alignment step also differs: every occurrence is aligned by \texttt{gpt-4.1-mini} (few-shot prompt in Appendix~\ref{app:prompt-align}, temperature 0, at most 20 output tokens), which is instructed to copy the translation of the term verbatim from the target segment, whereas the original pipeline calls the LLM only as a fallback for low-confidence spans of an embedding-based word aligner. We take this prompt from the organisers of the WMT25 terminology task \citep{semenov-etal-2025-findings}, with six in-context examples instead of their twenty. Applying the same aligner to every occurrence also keeps the extracted span from depending on which component handled the occurrence, which would otherwise be a source of spurious divergences in Step~2. Aligned spans and glossary terms are lemmatised with spaCy (\texttt{fr\_core\_news\_lg}) for the matching of Step~2.

Table~\ref{tab:variation-examples} illustrates the variation categories with which each attested source form is labelled relative to the head term of its concept. The label depends on how the form was detected: variants listed in the glossary, and the static variants that the \Concordancer\ generates from glossary entries, are classified by few-shot prompting \texttt{gpt-4.1}; head-term forms are \emph{no variation} by definition, and the forms that the \Concordancer\ detects in context mostly take the category of the formation pattern that produced them; a form derived from a glossary variant combines both (an acronym of an expansion is \emph{combined}). \citet{dahan-etal-2026-improving} validated an earlier version of this labelling (its own prompt, with \texttt{gpt-4.1-mini}) on 50 sampled forms annotated by a computational linguist: 84\% agreement on the category, near-perfect on single-category variants, with \emph{combined} variants under-detected (recall 11\%) but accounting for under 2\% of occurrences.

Term detection and the LLM alignment were validated manually on \ACL\ (Appendix~\ref{app:preproc-validation}): detection precision is 91.5\% and recall 95.5\%, and the alignment is correct for 78 of 80 randomly sampled occurrences, equally in the two prompting conditions.

\begin{table}[t]
\footnotesize
\centering
\resizebox{\columnwidth}{!}{%
\begin{tabular}{@{}lll@{}}
\toprule
\textbf{Category} & \textbf{Head term} & \textbf{Attested source form} \\
\midrule
no variation    & machine translation system & machine translation system \\
graphical       & machine translation system & MT system\\
morphosyntactic & machine translation system & system for machine translation \\
reduction       & machine translation system & translation system \\
expansion       & machine translation system  & automatic machine translation system \\
lexical         & machine translation system & automatic translation system \\
combined        & machine translation system & MT model \\
\bottomrule
\end{tabular}%
}
\caption{Source-side variation categories \citep{carcamo:hal-05442584}, illustrated on the running example of the introduction: each attested form is labelled relative to the head term of its concept.}
\label{tab:variation-examples}
\end{table}

Table~\ref{tab:worked-example} follows one concept of an \ACL\ document from its glossary entry to the verdicts; Table~\ref{tab:example-chain} then illustrates the consistency judge on real occurrences of a \STEP\ validation document.

\begin{table}[t]
\footnotesize
\centering
\setlength{\tabcolsep}{3pt}
\resizebox{\columnwidth}{!}{%
\begin{tabular}{@{}rlllc@{}}
\toprule
\textbf{Seg} & \textbf{Source form} & \textbf{MT output} & \textbf{Judgement} & \textbf{Verdict} \\
\midrule
\multicolumn{5}{@{}l@{}}{\emph{slab} (in glossary; variant \emph{downgoing slab} $\to$ \emph{plaque plongeante})} \\
6   & slab (1st mention) & plaque plongeante & first mention   & \VarJ \\
74  & slab               & plaque plongeante & document usage & \VarJ \\
495 & slab               & plaque plongeante & neutralisation & \VarNJ \\
\midrule
\multicolumn{5}{@{}l@{}}{\emph{subduction initiation} (out of glossary; expected translation \emph{initiation de la subduction})} \\
598 & subduction initiation & début de la subduction & paraphrase & \Atwo \\
\bottomrule
\end{tabular}%
}
\caption{Real occurrences from a \STEP\ validation document. The source alternates the head term \emph{slab} and its variant \emph{downgoing slab}; the system renders the variant consistently as \emph{plaque plongeante}. Rendering the head term the same way is judged legitimate at `first mention' and while it follows the `document usage' (\VarJ), but flagged once \emph{plaque plongeante} is the form reserved for the variant: the source distinction is erased (`neutralisation', \VarNJ). For the out-of-glossary concept \emph{subduction initiation}, whose reference is established from document usage, the paraphrase \emph{début de la subduction} is a terminological error (\Atwo).}
\label{tab:example-chain}
\end{table}

\section{Manual Validation of the Preprocessing}
\label{app:preproc-validation}

Term detection by the \Concordancer\ and the LLM alignment were validated manually on \ACL, the variation labelling having been validated in prior work (Appendix~\ref{app:pipeline}). All items were sampled, with a fixed random seed, from the aligned occurrences that \method\ scored in \S\ref{ssec:results-tal}: 11{,}122 occurrences in 34 documents, the same for the eight outputs since detection operates on the source text. One of the authors, a native French speaker fluent in English, answered one question per item (\emph{yes}, \emph{no} or \emph{uncertain}) without knowing which system or prompting condition had produced the translation; uncertain and unanswered items are left out. The questions concern the preprocessing only, not the quality of the translation.

For the \emph{precision of term detection}, 100 detected occurrences were shown in their source segment, with the glossary entry of the concept they were linked to, and the annotator judged whether the detected form refers to this concept, in its technical sense and with the right boundaries. Non-glossary variants were over-represented in this sample (half of the items, against a fifth of the occurrences), because they are rarer and, being derived beyond the glossary, more likely to be wrong; each answer is therefore weighted by the share of its detection type in the corpus. For the \emph{recall of term detection}, the annotator listed, in 40 source segments of at least six words (one per document, plus six), the occurrences of glossary concepts that no detection covered. For the \emph{alignment}, the annotator judged whether the proposed span is exactly the translation that the system wrote for the term, ignoring case, punctuation and a leading article; a wrong but correctly delimited translation counts as a correct alignment, and so does an empty span for an untranslated term. Thirteen aligned occurrences were drawn per output: ten at random among the occurrences whose span appears in the target segment, and three among those whose span is empty or absent from it (0.5 to 0.8\% of the occurrences of each output), where alignment errors are most likely.

\begin{table}[t]
\footnotesize
\centering
\begin{tabular}{@{}lrrr@{}}
\toprule
 & \textbf{Items} & \textbf{Correct} & \textbf{Rate} \\
\midrule
\multicolumn{4}{@{}l}{\emph{Term detection}} \\
\quad precision              & 96 & 82 & 91.5\%$^{*}$ \\
\quad recall                 & 66 & 63 & 95.5\% \\
\multicolumn{4}{@{}l}{\emph{Alignment}} \\
\quad \emph{baseline}        & 40 & 39 & 97.5\% \\
\quad \emph{base+terms}      & 40 & 39 & 97.5\% \\
\quad span empty or absent   & 22 & 18 & 81.8\% \\
\bottomrule
\end{tabular}
\caption{Manual validation of the preprocessing on \ACL. For detection precision and alignment, \emph{Items} counts the items answered \emph{yes} or \emph{no}, and \emph{Correct} those answered \emph{yes}. For recall, \emph{Items} counts the occurrences of glossary concepts in the 38 annotated segments, and \emph{Correct} those that the \Concordancer\ detected. $^{*}$Weighted by the share of each detection type in the corpus, since non-glossary variants were over-represented in the sample; unweighted, 82 of 96 (85.4\%). The \emph{baseline} and \emph{base+terms} rows sample, at random, the occurrences whose span appears in the target segment; the last row samples the others, whose span is empty or absent from the target segment (0.5 to 0.8\% of the occurrences of each output).}
\label{tab:preproc-validation}
\end{table}

Table~\ref{tab:preproc-validation} reports the results. Detection errors concentrate in non-glossary variants: seven of the fourteen come from the glossary variant \emph{base} (of \emph{morphological base}) being matched in \emph{based} and in compounds such as \emph{rule-based}, which accounts for 262 occurrences (2.4\% of the corpus), and most of the others are ordinary words linked to a concept, such as \emph{Table} in the caption ``Table~4'' (linked to \emph{confusion matrix}) or the verb \emph{split} in ``we will try to split those lines''. The three occurrences missed by the \Concordancer\ are \emph{metric}, \emph{pattern} and \emph{DBN5}; four other missed domain terms (e.g., \emph{decoder}) are absent from the glossary and are not counted. The two alignment errors on random occurrences extend the span to a neighbouring name (\emph{corpus} aligned to \emph{corpus Brown}). Among the spans empty or absent from the target segment, three of the four errors are words that the system did not translate, for which the aligner returned a translation anyway (\emph{instance} in \emph{for instance}), and the fourth is a defective copy of the translation (\emph{analogiede} for \emph{analogie}).

The samples locate only some of the errors, so they cannot be corrected one by one in the corpus. We therefore recomputed the \ACL\ scores of \S\ref{ssec:results-tal} from the verdicts of the published run, leaving out first the non-glossary variants (21.3\% of the occurrences, including the \emph{based} matches), then the occurrences whose span is empty or absent from the target segment. In both cases, the ranking of the four models within each prompting condition is unchanged, and \emph{base+terms} still scores better than \emph{baseline} for all four models.

\section{Dataset Construction Details}
\label{app:dataset-details}
The manual extraction behind the \biomqmterms\ glossary (\S\ref{sec:datasets}), performed independently of the terms' involvement in any translation error, produced 819 raw entries, deduplicated into 729 unique concepts. Crossing them with the bilingual MeSH thesaurus (30{,}915 concepts) covers 15.2\% of the terms by exact matching, and 31.7\% once alternative labels, inflected forms and graphical normalisation are admitted; the remaining 68.3\%, too specific or too compositional for MeSH, were translated manually. A complementary error-driven pass over the MQM Terminology error spans examined 132 candidate terms, of which 110 were retained. On \STEP, the 3{,}377 gold-labelled occurrences cover 261 concepts and 793 distinct source forms.

\section{A Typology of Terminological Errors}
\label{app:typology}
The error typology instantiated by the Q1 judge was developed for the human annotation of terminology errors in specialised machine translation \citep{carcamo:hal-05560658}. It builds on the MeLLANGE error typology \citep{kubler2008comparable} and on subsequent work on the translation of complex noun phrases in specialised texts \citep{kubler2022}, and its severity scale follows \citet{benard-etal-2024-etude}.

The typology is organised on three levels: a macro-category, a family (A to J), and a fine-grained code. Seven families describe errors in the translation of the term itself, independently of its other occurrences in the document, and yield the 23 fine-grained error codes of Table~\ref{tab:typology}: incorrect equivalent selection (A1--A7), constituent structure of complex terms (C1--C3), semantic relations between constituents (D1--D2), transposition (E1--E2), context-bound grammar and phraseology (F1--F4), content alteration (G1--G3), and target-language expression (H1--H2). Each fine-grained code carries a severity on the scale of \citet{benard-etal-2024-etude} (neutral, minor, major, critical; only 1--3 are attested in the typology table). Three families carry no severity: family~B (B1--B7) describes in-document terminological \emph{(in)consistency}: B1--B4 cover consistency facts about translations that are individually correct (the annotation guidelines stress that these cases « ne représentent pas vraiment une erreur » (\textsl{do not really correspond to errors})), B5--B7 combine incorrectness with instability, and its labels are never assigned alone, since consistency is only assessed at the document level; family~I marks the absence of error; family~J is an annotator-workflow tag. 

\method\ applies a fixed transformation to this typology before exposing it to the Q1 judge (Table~\ref{tab:label-mapping}). Family A is kept fine-grained; each of the families C to H is collapsed into a single grouped code carrying the gravity of its worst member; family B is excluded from Q1, since in-document inconsistency is exactly the question delegated to Q2 and keeping it in Q1 would count the same phenomenon twice; family I becomes the \texttt{no\_error} answer; family J is dropped as meaningless for an LLM judge. 
The 13 remaining codes are exposed to the judge as natural-language labels rather than opaque codes, which reduces reporting errors (mistyping \texttt{A6} for \texttt{A7} has no surface cue; \texttt{wrong\_component} vs \texttt{general\_word\_not\_term} does); the answer is remapped to the canonical code in post-processing, and any label outside the registry falls back to \texttt{no\_error}. The natural-language definitions and examples exposed to the judge for each of the 13 labels are those of the Q1 prompt, reproduced verbatim in Appendix~\ref{app:prompt-q1}.
The same transformation is applied to the human annotations of \STEP\ when validating the metric (\S\ref{ssec:step-protocol}), so that metric and human judgements live in the same 15-class space (the 13 codes, plus B reached through Q2, plus no-error).

\begin{table*}[t]
\footnotesize
\centering
\setlength{\tabcolsep}{4pt}
\newcolumntype{L}[1]{>{\raggedright\arraybackslash}p{#1}}
\newcolumntype{R}[1]{>{\raggedleft\arraybackslash}p{#1}}
\begin{tabular}{@{}L{2.7cm}L{2.0cm}L{2.2cm}R{0.7cm}L{2.7cm}L{4.2cm}@{}}
\toprule
\multicolumn{6}{@{}l@{}}{\textbf{Glossary entry}: \emph{semantic similarity} $\to$ \emph{similarité sémantique}; English variant \emph{semantic relatedness}; no French variant} \\
\midrule
\textbf{Source form (\Concordancer)} & \textbf{Variation category} & \textbf{In glossary} & \textbf{Occ.} & \textbf{Reference $R(s)$ (Step~1)} & \textbf{How the reference was fixed} \\
\midrule
semantic similarity  & no variation & head term            & 2  & similarité sémantique & glossary translation, taken as authoritative \\
similarity (incl.\ plural) & reduction & no                & 14 & similarité            & system's most frequent translation, validated by the error judge \\
semantic relatedness & lexical      & English variant only & 1  & parenté sémantique    & system's only translation, validated by the error judge \\
\bottomrule
\end{tabular}

\medskip
\begin{tabular}{@{}L{2.5cm}L{3.4cm}L{1.8cm}L{2.7cm}L{2.4cm}L{1.7cm}@{}}
\toprule
\textbf{Seg} & \textbf{Source form $\to$ MT output} & \textbf{Step 2: match?} & \textbf{Step 3: Q1} & \textbf{Step 4: Q2} & \textbf{Verdict} \\
\midrule
30, 42 & semantic similarity $\to$ similarité sémantique & yes & not called & not called & \Conforme\ ($\times$2) \\
33, 46, 47, 85, 87, 90$^\dagger$, 91, 93, 98, 103 ($\times$2), 110 & similarity $\to$ similarité(s) & yes & not called & not called & \Conforme\ ($\times$12) \\
79 & semantic relatedness $\to$ parenté sémantique & skipped (hapax) & no error, judged in absolute terms & not called (hapax) & \Conforme \\
90, 92 & similarity $\to$ similitudes & no: divergence & no error (one call, propagated) & gratuitous divergence & \VarNJ\ ($\times$2) \\
\bottomrule
\end{tabular}

\medskip
\begin{tabular}{@{}r p{0.47\textwidth} p{0.47\textwidth}@{}}
\toprule
\textbf{Seg} & \textbf{Source} (detected span in bold) & \textbf{MT output} (aligned span in bold) \\
\midrule
30  & \ldots\ strong link with the notion of \textbf{semantic similarity}. & \ldots\ lien étroit avec la notion de \textbf{similarité sémantique}. \\
33  & \ldots\ of a word according to the \textbf{similarity} with its semantic neighbors. & \ldots\ d'un mot en fonction de la \textbf{similarité} avec ses voisins sémantiques. \\
42  & \ldots\ based on the symmetry of \textbf{semantic similarity} relations. & \ldots\ sur la symétrie des relations de \textbf{similarité sémantique}. \\
46  & \ldots\ aim to calculate \textbf{similarities} between textual representations of word contexts. & \ldots\ visent à calculer les \textbf{similarités} entre les représentations textuelles des contextes \ldots \\
47  & Methods to calculate \textbf{similarities} from IR seem then relevant \ldots & Les méthodes pour calculer les \textbf{similarités} à partir de l'IR semblent alors \ldots \\
79  & \ldots\ our methods on semantic simlarity versus \textbf{semantic relatedness} relations. & \ldots\ nos méthodes sur les relations de similarité sémantique par rapport à celles de \textbf{parenté sémantique}. \\
85  & \ldots\ list of names ordered by decreasing \textbf{similarity}. & \ldots\ de noms classés par ordre de \textbf{similarité} décroissante. \\
87  & \ldots\ performance of different models of IR \textbf{similarities}. & \ldots\ les performances des différents modèles de \textbf{similarité} IR. \\
90$^\dagger$ & \ldots\ some IR \textbf{similarities} are quite inefficient including the TF alone or Hellinger \textbf{similarity}. & \ldots\ certaines \textbf{similitudes} IR sont assez inefficaces, notamment la similarité TF seule ou la \textbf{similarité} de Hellinger. \\
91  & \ldots\ since these \textbf{similarities} use very basic weights \ldots & \ldots\ puisque ces \textbf{similarités} utilisent des poids très basiques \ldots \\
92  & The \textbf{similarities} that include a notion of IDF \ldots & Les \textbf{similitudes} qui incluent une notion d'IDF \ldots \\
93  & Okapi BM25-based \textbf{similarities} offer good results. & Les \textbf{similarités} basées sur l'algorithme BM25 d'Okapi donnent \ldots \\
98  & Computing all the \textbf{similarities} between all pairs of words \ldots & Le calcul de toutes les \textbf{similarités} entre toutes les paires de mots \ldots \\
103 & The \textbf{similarity} between a word $w_i$ \ldots\ the same value as the \textbf{similarity} between the query $w_j$ \ldots & La \textbf{similarité} entre un mot $w_i$ \ldots\ la même valeur que la \textbf{similarité} entre la requête $w_j$ \ldots \\
110 & \ldots\ for giving a new \textbf{similarity} score in a simple way. & \ldots\ pour donner un nouveau score de \textbf{similarité} de manière simple. \\
\bottomrule
\end{tabular}
\caption{Worked example on one concept of an \ACL\ document translated by EuroLLM-22B-Instruct (\emph{baseline}). Top: the glossary entry, the forms detected by the \Concordancer\ with their variation category, and the reference translation fixed for each form in Step~1 of \S\ref{ssec:pipeline}; the glossary provides a French translation for the head term only, so the two other forms receive a pseudo-reference, the system's own translation validated by the error judge. Middle: the path of each occurrence through Steps~2 to~4 and the resulting verdict. Bottom: every segment involved. A match with the reference is labelled \Conforme\ without any LLM call. The hapax \emph{semantic relatedness} skips the match and is judged in absolute terms by the error judge, which accepts \emph{parenté sémantique}. The two occurrences rendered \emph{similitudes} diverge from the reference: the error judge finds no error (one call, its answer propagated to the identical translation), and the consistency judge classifies the switch as a `gratuitous divergence' from the document's usage (\VarNJ); segment 90 even keeps \emph{similarité} for \emph{Hellinger similarity} in the same sentence. $^\dagger$The \Concordancer\ detected only \emph{similarity} in \emph{Hellinger similarity}; the whole expression should have been identified as an expansion of the term and aligned to \emph{similarité de Hellinger}. In segment 79, the misspelt \emph{semantic simlarity} was not detected.}
\label{tab:worked-example}
\end{table*}

\onecolumn

{\footnotesize
\setlength{\tabcolsep}{4pt}
\renewcommand{\arraystretch}{1.05}
\setlength{\LTleft}{0pt}
\setlength{\LTright}{\fill}
\setlength{\LTcapwidth}{\textwidth}
\begin{longtable}{@{}l p{0.37\textwidth} p{0.33\textwidth} c c@{}}
\toprule
\textbf{Code} & \textbf{Definition} & \textbf{Example (source $\to$ *MT $\to$ expected)} & \textbf{Penalty} & \textbf{Support} \\
\midrule
\endfirsthead
\multicolumn{5}{@{}l}{\emph{Table~\ref{tab:typology} (continued)}}\\
\toprule
\textbf{Code} & \textbf{Definition} & \textbf{Example (source $\to$ *MT $\to$ expected)} & \textbf{Penalty} & \textbf{Support} \\
\midrule
\endhead
\multicolumn{5}{@{}l@{}}{\emph{A. Wrong equivalent selection}} \\
A1 & The translation is an attested term of the domain, but the equivalent of another term or concept: an inexact neighbour of the expected equivalent, which can also create spurious repetitions when the two concepts are close. & \emph{tremor} $\to$ *tremblement de terre $\to$ trémor & 1 & 111 \\
A2 & The terminological unit is replaced by a descriptive, general-language paraphrase, losing the precision of the established term. & \emph{lithospheric mantle} $\to$ *couche profonde sous la croûte terrestre $\to$ manteau lithosphérique & 1 & 20 \\
A3 & The system produces a formulation that is neither a recognised term of the domain nor a general-language expression: a literal creation. & \emph{earthquake swarm} $\to$ *sillage sismique $\to$ essaim sismique & 3 & 65 \\
A4 & The term is left in the source language although an established target-language equivalent exists. & \emph{megathrusts} $\to$ *megathrusts $\to$ mégachevauchements & 4 & 34 \\
A5 & A unit conventionally kept in the source language (proper name, acronym, nomenclature symbol) is translated; for acronyms, the letters may also be reworked or reordered. & \emph{MORB} $\to$ *BDRM $\to$ MORB & 4 & 23 \\
A6 & Some constituents of a complex term are translated correctly but others are not, yielding a partially correct hybrid term. & \emph{fault patches} $\to$ *taches de faille $\to$ zones de faille & 4 & 95 \\
A7 & A single general-language word replaces the specialised term; unlike A2, one lexical unit rather than a paraphrase, creating referential ambiguity in specialised discourse. & \emph{repeaters} $\to$ *répétiteurs $\to$ séismes répétitifs & 3 & 23 \\
\multicolumn{5}{@{}l@{}}{\emph{B. In-document terminological (in)consistency (no severity; delegated to the consistency judge)}} \\
B1 & The same base term receives different, individually correct translations across the document. & \emph{earthquake rupture} $\to$ rupture sismique / ruptures de séismes & -- & -- \\
B2 & The same variant receives different, individually correct translations across the document. & \emph{fault creep} $\to$ fluage asismique / fluage de la faille / fluage de faille & -- & -- \\
B3 & A base term and one of its variants receive the same translation, erasing the distinction between them. & \emph{seismic hazard} (variant) / \emph{earthquake hazard} (base term) $\to$ risque sismique & -- & -- \\
B4 & Different variants of the same base term receive the same translation, erasing their nuances. & \emph{dynamic fault slip} / \emph{dynamic slip} $\to$ glissement dynamique & -- & -- \\
B5 & The base term is translated both incorrectly and inconsistently from one occurrence to the next. & \emph{continental forearc} $\to$ *marge continentale / *zone de subduction / *zone de l'avant-arc continental & -- & -- \\
B6 & A variant is translated both incorrectly and inconsistently from one occurrence to the next. & \emph{down-dip} $\to$ *encaissée / *en profondeur / *dans la direction aval & -- & -- \\
B7 & The base term and its variant both receive the same incorrect translation, combining neutralisation with incorrectness. & \emph{Zapotal Formation} / \emph{Zapotal Fm} $\to$ *Formation de Zapotal & -- & -- \\
\multicolumn{5}{@{}l@{}}{\emph{C. Structure of complex terms}} \\
C1 & The head noun of the complex term or noun phrase is misidentified; the head fixing the base category of the concept, the designated concept changes. & \emph{subduction-zone serpentinites} $\to$ *zones de subduction constituées de serpentinite $\to$ serpentinites en zone de subduction & 4 & 10 \\
C2 & An adjective or other modifier is attached to the wrong constituent, altering the qualification relations within the term or phrase. & \emph{fast earthquake rupture} $\to$ *rupture rapide de séisme $\to$ rupture sismique rapide & 4 & 11 \\
C3 & The constituents are ordered against the conventions of the target language, yielding a structure alien to domain usage. & \emph{slow earthquake} $\to$ *lent séisme $\to$ séisme lent & 4 & 9 \\
\multicolumn{5}{@{}l@{}}{\emph{D. Semantic relations}} \\
D1 & The semantic relations between constituents, often implicit in scientific English, are rendered incorrectly where the target language requires them to be explicit (preposition or paraphrase). & \emph{fast and slow events} $\to$ *événements à la fois rapides et lents $\to$ événements rapides et lents & 4 & 3 \\
D2 & An element shared (factorised) across coordinated constituents in the source is not distributed to all conjuncts in the target. & \emph{pressure-solution and dislocation creep} $\to$ *pression-solution et le glissement par dislocation $\to$ fluage par dissolution-précipitation et par dislocation & 4 & 0 \\
\multicolumn{5}{@{}l@{}}{\emph{E. Transposition}} \\
E1 & The compact, synthetic source structure is kept as is, without the explicitation (preposition, article, unfolding) that the analytic target structure requires. & \emph{strike-slip and subduction thrust faults} $\to$ *failles de décrochement et de chevauchement de subduction $\to$ failles de décrochement et de chevauchement dans les zones de subduction & 3 & 13 \\
E2 & The source syntax is calqued although the words themselves are attested, producing unnatural target text (e.g., an English plural mark kept on an acronym, French acronyms being invariable). & \emph{SSEs} $\to$ *SSEs $\to$ SSE & 1 & 39 \\
\multicolumn{5}{@{}l@{}}{\emph{F. Grammar / phraseology in context}} \\
F1 & Inappropriate determiner, or incorrect gender or number agreement, within the term or between the term and its context. & \emph{alkaline basalt} $\to$ *la basalte alcalin $\to$ le basalte alcalin & 1 & 132 \\
F2 & A preposition required by the target syntactic structure is missing or wrong. & \emph{slip behavior} $\to$ *comportement en glissement $\to$ comportement de glissement & 3 & 71 \\
F3 & A grammatically correct combination that does not match the established phraseology of the domain. & \emph{magma rises up the conduit} $\to$ *le magma monte vers le haut dans le conduit $\to$ le magma remonte par le conduit & 3 & 0 \\
F4 & An incorrect spelling, notably in terms borrowed or adapted from the source language. & \emph{aseismic slip} $\to$ *glissement aseismique $\to$ glissement asismique & 3 & 16 \\
\multicolumn{5}{@{}l@{}}{\emph{G. Content alteration}} \\
G1 & Constituents absent from the source term are added, potentially altering the meaning or introducing unintended nuances. & \emph{fault creep} $\to$ *glissement lent de la faille $\to$ fluage de faille & 4 & 10 \\
G2 & Constituents of the term, or the term entirely, are omitted, losing information that affects the meaning. & \emph{seismicity bursts} $\to$ *séismes $\to$ vagues de séismes & 4 & 56 \\
G3 & The output bears no relation to the source content or is incoherent with the term to be translated: an unintelligible translation or hallucination. & \emph{earthquake ruptures} $\to$ *ruptures d'éruptions $\to$ ruptures sismiques & 4 & 161 \\
\multicolumn{5}{@{}l@{}}{\emph{H. Target-language expression}} \\
H1 & The relative weight and hierarchy of the constituents, notably their punctuation, is not respected in the target text. & -- & 3 & 0 \\
H2 & A semantically correct formulation whose register or style is inappropriate for specialised discourse. & \emph{earthquake ruptures} $\to$ *ruptures de tremblements de terre $\to$ ruptures sismiques & 1 & 16 \\
\bottomrule
\noalign{\smallskip}
\caption{The fine-grained codes of the typology \citep{carcamo:hal-05560658} as implemented in \method: the 23 error codes and the seven consistency codes of family B, with definitions adapted from the annotation guidelines and a geoscience example each (*: erroneous MT output; for error codes, the last element is the expected form). The default penalty follows the fixed 1/3/4 mapping of the gravities assigned by the typology's authors on the scale of \citet{benard-etal-2024-etude}; support is the number of gold occurrences carrying the code among the 3{,}373 evaluated \STEP\ occurrences (precedence-reduced labels). Family B carries no severity and is never assigned alone: it is the question delegated to the consistency judge (Q2), and the precedence reduction of \S\ref{ssec:step-protocol} collapses pure-consistency annotations into a single B class (101 occurrences), so no per-code support is reported.}
\label{tab:typology}
\end{longtable}}
\twocolumn

\begin{table}[t]
\footnotesize
\centering
\setlength{\tabcolsep}{4pt}
\renewcommand{\arraystretch}{0.95}

\resizebox{\columnwidth}{!}{%
\begin{tabular}{@{}llc@{}}
\toprule
\textbf{Prompt label} & \textbf{Code} & \textbf{Covers} \\
\midrule
\texttt{wrong\_related\_term}        & \Aone   & A1 \\
\texttt{paraphrase\_not\_term}       & \Atwo   & A2 \\
\texttt{invented\_calque}            & \Athree & A3 \\
\texttt{left\_untranslated}          & \Afour  & A4 \\
\texttt{dnt\_violated}               & \Afive  & A5 \\
\texttt{wrong\_component}            & \Asix   & A6 \\
\texttt{general\_word\_not\_term}    & \Aseven & A7 \\
\texttt{misparsed\_structure}        & \textbf{C} & C1--C3 \\
\texttt{wrong\_constituent\_relation} & \textbf{D} & D1--D2 \\
\texttt{failed\_transposition}       & \textbf{E} & E1--E2 \\
\texttt{grammar\_phraseology}        & \textbf{F} & F1--F4 \\
\texttt{content\_altered}            & \textbf{G} & G1--G3 \\
\texttt{defective\_register\_form}   & \textbf{H} & H1--H2 \\
\texttt{no\_error}                   & --      & I \\
\bottomrule
\end{tabular}%
}
\caption{Transformation of the error typology for the Q1 judge: family A stays fine-grained, families C--H are collapsed into one grouped code each, family B is delegated to Q2, family J is dropped. The judge answers with the natural-language label; the canonical code is restored in post-processing.}
\label{tab:label-mapping}
\end{table}

\subsection{Why the Compact Typology}
\label{app:typology-comparison}

The choice of the compact typology over the full 23-code version rests on a controlled comparison on \STEP: two runs of the metric identical in every respect, including references verified identical occurrence by occurrence, except the typology exposed to the Q1 judge, with every occurrence routed through the exhaustive judging path. The unbiased comparison between the two is binary error detection, since the binarised gold is the same for both runs (1{,}019 erroneous and 2{,}354 error-free occurrences); Table~\ref{tab:typology-comparison} shows that the compact typology dominates on every measure, with 2.8 points more accuracy, 5.3 points more error precision, 78 fewer false positives and 17 fewer false negatives. Over-segmenting the answer space does not only degrade category choice: it degrades the binary decision itself, pushing the judge out of \texttt{no\_error} more often and wrongly. The multi-class views live in different label spaces and are not directly comparable, but two observations are robust: the full typology leaves seven attested categories at exactly zero F1 (A5, C1, C2, D1, E1, F2, G1), and the family-A diagnostics, computed identically in both spaces, are of the same order (detection of some A code on gold-A occurrences at 0.590 for the compact typology against 0.563; exact A code among those detected at 0.521 against 0.536), so the compaction costs nothing on the fine-grained terminological core. The full prompt is also about twice as expensive (about 6{,}700 input tokens per Q1 call against 3{,}100) and overruns the output budget where the compact prompt never does.

The compaction is deliberately asymmetric. Family A keeps its seven fine codes because distinguishing the mechanisms of wrong equivalent selection is precisely what the metric must explain; families C to H describe errors that affect terms without being terminology errors proper, so the metric only needs to know which family an occurrence fails under. The distributional picture supports the same asymmetry: taken together, families C to H account for 547 of the 1{,}019 erroneous occurrences and are in no way marginal, but their sixteen fine codes are individually rare (median support 12; eleven of the sixteen at 16 occurrences or fewer; D2, F3 and H1 unattested; only F1 and G3 above one hundred; Table~\ref{tab:typology}), too rare for the judge to be evaluated reliably on them or for the gold to discriminate them, whereas the grouped families recover workable supports (C~30, E~52, F~219, G~227). This observation is consistent with prior findings that fine-grained MQM annotation is much harder for LLMs than coarser quality judgements \citep{fernandes-etal-2023-devil}, that in-context classification degrades over large label spaces \citep{milios-etal-2023-many-labels}, and with the practice of predicting spans and severities rather than the full MQM taxonomy \citep{guerreiro-etal-2024-xcomet}; to our knowledge, no prior study quantifies the effect of typology size on an LLM judge all else being equal, which this controlled comparison provides. Its limits are those of the setting: one corpus, one MT output, one judge model, and a consensus gold reduced from multi-label annotations by a fixed precedence.

\begin{table}[t]
\footnotesize
\centering
\setlength{\tabcolsep}{4pt}

\begin{tabular}{@{}lcc@{}}
\toprule
\textbf{Binary error detection} & \textbf{Compact} & \textbf{Full} \\
\midrule
Accuracy            & \textbf{0.775} & 0.747 \\
Macro-F1            & \textbf{0.730} & 0.701 \\
Balanced accuracy   & \textbf{0.728} & 0.703 \\
Precision (error class)   & \textbf{0.632} & 0.579 \\
Recall (error class)      & \textbf{0.609} & 0.593 \\
F1 (error class)          & \textbf{0.620} & 0.586 \\
False positives     & \textbf{362}   & 440 \\
False negatives     & \textbf{398}   & 415 \\
\bottomrule
\end{tabular}
\caption{Controlled comparison of the compact (13-code) and full (23-code) typologies on \STEP: binary error detection against the same binarised gold (3{,}373 occurrences; 1{,}019 error, 2{,}354 no-error). The two runs differ only in the typology exposed to the Q1 judge.}
\label{tab:typology-comparison}
\end{table}

\section{\STEP\ Confusion Matrix}
\label{app:confusion}

Table~\ref{tab:confusion} gives the full confusion matrix behind the agreement figures of \S\ref{ssec:results-step}, with every occurrence judged through the exhaustive Q1$\to$Q2 path.

\begin{table*}[t]
\scriptsize
\centering
\setlength{\tabcolsep}{3pt}

\resizebox{\textwidth}{!}{%
\begin{tabular}{@{}lrrrrrrrrrrrrrrrr@{}}
\toprule
\textbf{Gold} $\backslash$ \textbf{verdict} & A1 & A2 & A3 & A4 & A5 & A6 & A7 & C & D & E & F & G & H & no-error & B & total \\
\midrule
A1 & \textbf{74} & 2 & 0 & 0 & 0 & 10 & 7 & 0 & 0 & 0 & 3 & 9 & 0 & 4 & 2 & 111 \\
A2 & 2 & \textbf{9} & 0 & 0 & 0 & 0 & 0 & 0 & 0 & 1 & 0 & 0 & 0 & 7 & 1 & 20 \\
A3 & 4 & 4 & \textbf{4} & 0 & 0 & 0 & 1 & 5 & 0 & 0 & 5 & 0 & 0 & 40 & 2 & 65 \\
A4 & 0 & 0 & 12 & \textbf{8} & 0 & 4 & 0 & 0 & 0 & 1 & 0 & 0 & 0 & 9 & 0 & 34 \\
A5 & 2 & 0 & 2 & 0 & \textbf{6} & 1 & 0 & 1 & 0 & 0 & 0 & 0 & 0 & 9 & 2 & 23 \\
A6 & 34 & 7 & 3 & 0 & 0 & \textbf{6} & 7 & 0 & 2 & 0 & 1 & 0 & 0 & 28 & 7 & 95 \\
A7 & 1 & 1 & 0 & 0 & 0 & 1 & \textbf{7} & 1 & 0 & 0 & 0 & 0 & 0 & 5 & 7 & 23 \\
C & 1 & 3 & 2 & 0 & 0 & 2 & 0 & \textbf{4} & 0 & 1 & 0 & 6 & 1 & 9 & 1 & 30 \\
D & 0 & 0 & 0 & 0 & 0 & 0 & 0 & 1 & \textbf{0} & 0 & 0 & 1 & 0 & 1 & 0 & 3 \\
E & 0 & 0 & 0 & 8 & 0 & 0 & 0 & 2 & 0 & \textbf{9} & 1 & 0 & 0 & 28 & 4 & 52 \\
F & 0 & 0 & 0 & 6 & 0 & 13 & 0 & 0 & 1 & 10 & \textbf{17} & 2 & 0 & 159 & 11 & 219 \\
G & 126 & 22 & 8 & 2 & 2 & 7 & 11 & 0 & 0 & 1 & 0 & \textbf{32} & 0 & 12 & 4 & 227 \\
H & 2 & 8 & 0 & 0 & 0 & 0 & 1 & 0 & 0 & 0 & 0 & 0 & \textbf{0} & 5 & 0 & 16 \\
no-error & 35 & 52 & 8 & 34 & 3 & 26 & 16 & 6 & 3 & 17 & 40 & 16 & 2 & \textbf{1992} & 104 & 2354 \\
B & 1 & 1 & 0 & 0 & 0 & 4 & 0 & 0 & 0 & 0 & 2 & 6 & 1 & 82 & \textbf{4} & 101 \\
\midrule
total & 282 & 109 & 39 & 58 & 11 & 74 & 50 & 20 & 6 & 40 & 69 & 72 & 4 & 2390 & 149 & 3373 \\
\bottomrule
\end{tabular}%
}
\caption{Confusion matrix of \method's verdicts against the \STEP\ gold labels (3{,}373 scored occurrences; gold labels in rows, verdicts in columns, diagonal in bold). On the verdict side, the no-error column aggregates \Conforme\ and \VarJ, and the B column is \VarNJ\ (mapping of \S\ref{ssec:step-protocol}); row totals are the gold supports of Table~\ref{tab:step-results}.}
\label{tab:confusion}
\end{table*}

\section{Macro Aggregation Levels}
\label{app:macro-agg}

Tables~\ref{tab:biomqm-form} and~\ref{tab:biomqm-concept} complete Table~\ref{tab:biomqm-occ}, reported under the micro-average (occurrence) aggregation, with the two macro-aggregations of \S\ref{ssec:aggregation}, per source form and per concept. The conclusions are unchanged: \method\ ranks first in SPA at both levels, ahead of every comparison family.

\begin{table}[t]
\footnotesize
\centering
\setlength{\tabcolsep}{3.5pt}

\resizebox{\columnwidth}{!}{%
\begin{tabular}{@{}lcc@{}}
\toprule
\textbf{Metric} & \textbf{SPA} & \textbf{\acceq} (doc) \\
\midrule
\multicolumn{3}{@{}l}{\emph{Judged divergences (this work)}} \\
\textbf{\method}            & \textbf{0.871} & 0.556 \\
\multicolumn{3}{@{}l}{\emph{Divergence QE}} \\
\quad \textsc{Gemba}$_\text{div}$        & 0.815 & 0.524 \\
\quad MetricX-24$_\text{div}$            & 0.797 & 0.536 \\
\quad CometKiwi$_\text{div}$             & 0.804 & 0.552 \\
\multicolumn{3}{@{}l}{\emph{Glossary conformity}} \\
\quad first-translation fallback & 0.766 & 0.458 \\
\quad majority fallback     & 0.758 & 0.454 \\
\multicolumn{3}{@{}l}{\emph{Terminology-filtered QE}} \\
\quad \textsc{Gemba-MQM} (any mention) & 0.730 & 0.209 \\
\quad \textsc{Gemba-MQM} (category)    & 0.646 & 0.183 \\
\multicolumn{3}{@{}l}{\emph{General QE}} \\
\quad \textsc{Gemba}$_\text{gen}$        & 0.721 & 0.530 \\
\quad MetricX-24$_\text{gen}$            & 0.728 & 0.568 \\
\quad CometKiwi$_\text{gen}$             & 0.747 & 0.549 \\
\bottomrule
\end{tabular}%
}
\caption{Meta-evaluation on \biomqmterms, \textbf{source-form} (macro) aggregation. General-QE rows are unchanged across aggregations (only the human side varies for them).}
\label{tab:biomqm-form}
\end{table}

\begin{table}[t]
\footnotesize
\centering
\setlength{\tabcolsep}{3.5pt}
\resizebox{\columnwidth}{!}{%
\begin{tabular}{@{}lcc@{}}
\toprule
\textbf{Metric} & \textbf{SPA} & \textbf{\acceq} (doc) \\
\midrule
\multicolumn{3}{@{}l}{\emph{Judged divergences (this work)}} \\
\textbf{\method}            & \textbf{0.867} & 0.562 \\
\multicolumn{3}{@{}l}{\emph{Divergence QE}} \\
\quad \textsc{Gemba}$_\text{div}$        & 0.834 & 0.518 \\
\quad MetricX-24$_\text{div}$            & 0.811 & 0.535 \\
\quad CometKiwi$_\text{div}$             & 0.823 & 0.549 \\
\multicolumn{3}{@{}l}{\emph{Glossary conformity}} \\
\quad first-translation fallback & 0.780 & 0.447 \\
\quad majority fallback     & 0.771 & 0.443 \\
\multicolumn{3}{@{}l}{\emph{Terminology-filtered QE}} \\
\quad \textsc{Gemba-MQM} (any mention) & 0.725 & 0.204 \\
\quad \textsc{Gemba-MQM} (category)    & 0.641 & 0.178 \\
\multicolumn{3}{@{}l}{\emph{General QE}} \\
\quad \textsc{Gemba}$_\text{gen}$        & 0.736 & 0.527 \\
\quad MetricX-24$_\text{gen}$            & 0.744 & 0.566 \\
\quad CometKiwi$_\text{gen}$             & 0.766 & 0.556 \\
\bottomrule
\end{tabular}%
}
\caption{Meta-evaluation on \biomqmterms, \textbf{concept} (macro) aggregation.}
\label{tab:biomqm-concept}
\end{table}

\section{Penalty Scale and Sensitivity}
\label{app:penalties}
This appendix details the penalty scale behind the document score of \S\ref{ssec:results-biomqm} and its sensitivity analysis.

Like any typology-based metric, e.g. the Critical/Major/Minor weighting of MQM, (\S\ref{ssec:evaluation-methods}), the document score rests on a penalty scale. The fine-grained penalties are not tuned: they follow the gravities assigned to each code by the typology's authors, through the fixed mapping of gravities 1/2/3 to weights 1/3/4 (Appendix~\ref{app:typology}). The 1/3/4 spacing compresses the Minor/Major/Critical scale of MQM penalties (1/5/10): the decisive gap separates near-misses from established errors, while gravities 2 and 3 both signal an unusable translation of the term; the compression also bounds the dynamic range, preventing a few critical codes from dominating a document score. A grouped verdict asserts an error of the family but does not identify which sub-code applies, and the sub-codes of a family differ in gravity; each grouped family, C through H, is therefore scored at its most lenient sub-code, a single benefit-of-the-doubt rule applied uniformly, in favour of the evaluated translation. The uncertainty lies in the judge's granularity, not in the translation, so the rule is a conservative bound rather than an indulgence: the penalty never exceeds what the verdict certainly establishes. D, with only 31 of the 13{,}200 occurrences, is too rare for its weight to have any measurable effect.

The six grouped-family penalties (C--H) are the only quantities this rule sets. To test how sensitive the meta-evaluation is to them, we re-scored the saved verdicts under every assignment of weights 1--4 to these six families ($4^{6}=4{,}096$ configurations; A~weights and the \VarNJ\ penalty at their defaults; macro-average per lemmatised source form), at no LLM cost, since the judges emit labels and the score is a pure function of the penalties applied to them. The meta-evaluation responds in a single, interpretable direction: averaged over the grid, raising the weight of family F (grammar around the term) costs six SPA points (0.833 to 0.774), G (content alteration) three and a half, H (target-language expression) one and a half, E half a point, while C and D (term structure, semantic relations) have no measurable effect. These are the same peripheral families that the confusion analysis of \S\ref{ssec:results-step} shows that the judge handles least reliably; inflating their weight adds non-terminological noise into the score. The minimum rule, adopted on benefit-of-the-doubt grounds alone, thus receives independent support: it places the scale (Table~\ref{tab:verdicts}) in the $F{=}G{=}H{=}1$ region, whose 64 configurations keep SPA within 2.5 points of one another (0.846--0.871) and above every comparison metric at that aggregation (Table~\ref{tab:biomqm-form}), whatever the C, D and E weights, and re-scoring the same 64 scales at the micro aggregation of Table~\ref{tab:biomqm-occ} keeps every one of them above every comparison metric as well; the \VarNJ\ penalty is second-order (varying it from 0.5 to 1.2 moves mean SPA by a tenth of a point on the development configuration).

\section{Meta-Evaluation without Hapax}
\label{app:nohapax}

Tables~\ref{tab:nohapax-occ}, \ref{tab:nohapax-form} and~\ref{tab:nohapax-concept} repeat the meta-evaluation of \S\ref{ssec:results-biomqm} with single-occurrence chains excluded from both the metrics and the human scores (48 documents, 275 segments). Two caveats: the sentence-level rows, being term-agnostic, keep their full-document scores against the complete human ground truth (50 documents), and are repeated unchanged; and the variant removes exactly the occurrences that only a judge can evaluate, so it is not comparable to the main tables. On this reduced support the SPA differences between systems are all within statistical noise (paired permutation tests, $p \geq 0.17$ for every pair involving \method, $k{=}2{,}000$): the apparent reordering, with the glossary-conformity baselines nominally ahead, is not significant. \method\ remains significantly better than the five sentence-level rows on segment-level \acceq ($p < 0.05$) and keeps the best document-level \acceq\ of the terminology-aware families. 

\begin{table}[t]
\footnotesize
\centering
\setlength{\tabcolsep}{3.5pt}
\resizebox{\columnwidth}{!}{%
\begin{tabular}{@{}lccc@{}}
\toprule
\textbf{Metric} & \textbf{SPA} & \textbf{\acceq} (seg) & \textbf{\acceq} (doc) \\
\midrule
\multicolumn{4}{@{}l}{\emph{Judged divergences (this work)}} \\
\method                     & 0.750 & 0.679 & 0.464 \\
\multicolumn{4}{@{}l}{\emph{Divergence QE}} \\
\quad \textsc{Gemba}$_\text{div}$        & 0.725 & 0.671 & 0.393 \\
\quad MetricX-24$_\text{div}$            & 0.752 & 0.675 & 0.451 \\
\quad CometKiwi$_\text{div}$             & 0.752 & 0.678 & 0.458 \\
\multicolumn{4}{@{}l}{\emph{Glossary conformity}} \\
\quad first-translation fallback & \textbf{0.778} & 0.676 & 0.406 \\
\quad majority fallback     & 0.770 & \textbf{0.680} & 0.403 \\
\multicolumn{4}{@{}l}{\emph{Terminology-filtered QE (50 docs)}} \\
\quad \textsc{Gemba-MQM} (any mention) & 0.775 & 0.653 & 0.219 \\
\quad \textsc{Gemba-MQM} (category)    & 0.689 & 0.653 & 0.193 \\
\multicolumn{4}{@{}l}{\emph{General QE (50 docs)}} \\
\quad \textsc{Gemba}$_\text{gen}$        & 0.736 & 0.650 & 0.533 \\
\quad MetricX-24$_\text{gen}$            & 0.736 & 0.650 & 0.566 \\
\quad CometKiwi$_\text{gen}$             & 0.729 & 0.651 & 0.552 \\
\bottomrule
\end{tabular}%
}
\caption{No-hapax meta-evaluation on \biomqmterms, micro-average (occurrence) aggregation (48 documents, 275 segments, except term-agnostic sentence-level rows: 50 documents, full human ground truth).}
\label{tab:nohapax-occ}
\end{table}

\begin{table}[t]
\footnotesize
\centering
\setlength{\tabcolsep}{3.5pt}
\resizebox{\columnwidth}{!}{%
\begin{tabular}{@{}lcc@{}}
\toprule
\textbf{Metric} & \textbf{SPA} & \textbf{\acceq} (doc) \\
\midrule
\multicolumn{3}{@{}l}{\emph{Judged divergences (this work)}} \\
\method                     & 0.741 & 0.461 \\
\multicolumn{3}{@{}l}{\emph{Divergence QE}} \\
\quad \textsc{Gemba}$_\text{div}$        & 0.708 & 0.388 \\
\quad MetricX-24$_\text{div}$            & 0.750 & 0.452 \\
\quad CometKiwi$_\text{div}$             & 0.735 & 0.450 \\
\multicolumn{3}{@{}l}{\emph{Glossary conformity}} \\
\quad first-translation fallback & \textbf{0.785} & 0.400 \\
\quad majority fallback     & 0.776 & 0.396 \\
\multicolumn{3}{@{}l}{\emph{Terminology-filtered QE (50 docs)}} \\
\quad \textsc{Gemba-MQM} (any mention) & 0.730 & 0.209 \\
\quad \textsc{Gemba-MQM} (category)    & 0.646 & 0.183 \\
\multicolumn{3}{@{}l}{\emph{General QE (50 docs)}} \\
\quad \textsc{Gemba}$_\text{gen}$        & 0.721 & 0.530 \\
\quad MetricX-24$_\text{gen}$            & 0.728 & 0.568 \\
\quad CometKiwi$_\text{gen}$             & 0.747 & 0.549 \\
\bottomrule
\end{tabular}%
}
\caption{No-hapax meta-evaluation, source-form aggregation.}
\label{tab:nohapax-form}
\end{table}

\begin{table}[t]
\footnotesize
\centering
\setlength{\tabcolsep}{3.5pt}
\resizebox{\columnwidth}{!}{%
\begin{tabular}{@{}lcc@{}}
\toprule
\textbf{Metric} & \textbf{SPA} & \textbf{\acceq} (doc) \\
\midrule
\multicolumn{3}{@{}l}{\emph{Judged divergences (this work)}} \\
\method                     & 0.731 & 0.464 \\
\multicolumn{3}{@{}l}{\emph{Divergence QE}} \\
\quad \textsc{Gemba}$_\text{div}$        & 0.717 & 0.384 \\
\quad MetricX-24$_\text{div}$            & 0.754 & 0.450 \\
\quad CometKiwi$_\text{div}$             & 0.744 & 0.450 \\
\multicolumn{3}{@{}l}{\emph{Glossary conformity}} \\
\quad first-translation fallback & \textbf{0.794} & 0.401 \\
\quad majority fallback     & 0.785 & 0.398 \\
\multicolumn{3}{@{}l}{\emph{Terminology-filtered QE (50 docs)}} \\
\quad \textsc{Gemba-MQM} (any mention) & 0.725 & 0.204 \\
\quad \textsc{Gemba-MQM} (category)    & 0.641 & 0.178 \\
\multicolumn{3}{@{}l}{\emph{General QE (50 docs)}} \\
\quad \textsc{Gemba}$_\text{gen}$        & 0.736 & 0.527 \\
\quad MetricX-24$_\text{gen}$            & 0.744 & 0.566 \\
\quad CometKiwi$_\text{gen}$             & 0.766 & 0.556 \\
\bottomrule
\end{tabular}%
}
\caption{No-hapax meta-evaluation, concept-level aggregation.}
\label{tab:nohapax-concept}
\end{table}

\section{Exhaustive Judging Path on \iwslt}
\label{app:exhaustive}

On \iwslt\ we re-ran the eight outputs through the exhaustive judging path: no deterministic shortcut, every occurrence submitted to Q1 and, when error-free, to Q2. Table~\ref{tab:exhaustive} compares this path with the default configuration: the ranking is reproduced (Kendall $\tau = 0.786$; 25 of 28 system pairs concordant, the three discordances all involving near-ties, and every \emph{base+terms} system still ahead of every \emph{baseline} one) while requiring 2.55 times more unique LLM calls. The deterministic shortcuts and the Q1 propagation thus buy a substantial cost reduction without affecting the conclusions.

\begin{table}[t]
\footnotesize
\centering
\setlength{\tabcolsep}{4pt}
\begin{tabular}{@{}lcc@{}}
\toprule
 & \textbf{Default} & \textbf{Exhaustive} \\
\midrule
Unique LLM judgements & 6{,}525 & 16{,}612 \\
Approximate cost (USD) & 6.8 & 15.6 \\
Kendall $\tau$ between rankings & \multicolumn{2}{c}{0.786 (25/28 pairs)} \\
\emph{base+terms} ahead of \emph{baseline} & 16/16 & 16/16 \\
\bottomrule
\end{tabular}
\caption{Default configuration (deterministic shortcuts, Q1 propagation) vs.\ exhaustive judging path on the eight \iwslt\ outputs, same judge and prompts. The three discordant pairs all involve near-ties (score gap $\leq 0.006$ in one of the two configurations).}
\label{tab:exhaustive}
\end{table}

\section{Consistency Labels: Grounding and Distribution}
\label{app:q2-labels}

The ten labels of the consistency judge operationalise motivations for terminological variation documented in terminology and translation studies. Among the seven coherent labels, `first mention' is the base case: no translation is established yet, so none can be inconsistent. `Explicitation' follows the translation procedure of that name \citep{vinay-darbelnet-1972} and the cognitive function of variation, by which writers vary a term to explain a concept better \citep{pecman:hal-01232653}. `Avoid repetition' reflects the stylistic avoidance of repetition, marked in French \citep{vinay-darbelnet-1972}. `Register' and `facet' follow \citet{bowker-1998}, who shows that variation in expert discourse is motivated by the audience and purpose of the text and by the dimension of the concept that the context foregrounds, a cognitive function also described by \citet{pecman:hal-01232653}. `Document usage' and `synonym merge' encode the consistency norm itself: repeating the rendering established in the document is coherent, and two source forms that differ only by syntactic reordering (\emph{colour flatbed scanner} vs \emph{flatbed colour scanner} in \citealp{bowker-1998}; morphosyntactic variants in \citealp{carcamo:hal-05442584}) may share one translation. The three inconsistency labels mirror the consistency tradition of \S\ref{sec:related}: `synonym inconsistency' and `gratuitous divergence' penalise a switch away from an established rendering without such a motivation, and `neutralisation' the erasure of a source-side distinction \citep{bowker-hawkins-2006}.

Table~\ref{tab:q2-labels} details, per system and prompting condition, the distribution of the consistency-judge labels on \ACL\ behind the analysis of \S\ref{sec:analysis}.

\begin{table}[t]
\footnotesize
\centering
\setlength{\tabcolsep}{1.5pt}
\resizebox{\columnwidth}{!}{%
\begin{tabular}{@{}lcccccccc@{}}
\toprule
& \multicolumn{4}{c}{\emph{baseline}} & \multicolumn{4}{c}{\emph{base+terms}} \\
\cmidrule(lr){2-5}\cmidrule(lr){6-9}
\textbf{Q2 label (\%)} & Ll. & Qw. & E9 & E22 & Ll. & Qw. & E9 & E22 \\
\midrule
Document usage    & 79.5 & 76.5 & 79.2 & 80.1 & 78.8 & 77.2 & 79.8 & 81.0 \\
First mention     & 6.0 & 6.7 & 6.6 & 6.1 & 6.6 & 7.0 & 7.3 & 6.0 \\
Synonym merge     & 2.2 & 3.0 & 2.0 & 2.1 & 1.7 & 1.8 & 1.8 & 1.4 \\
Other coherent    & 0.1 & 0.2 & 0.3 & 0.3 & 0.3 & 0.2 & 0.2 & 0.4 \\
Neutralisation    & 0.8 & 1.8 & 1.2 & 1.6 & 2.1 & 1.8 & 1.5 & 1.6 \\
Synonym inconsist. & 3.7 & 3.3 & 3.0 & 3.2 & 2.0 & 2.4 & 2.9 & 2.3 \\
Grat. divergence  & 7.8 & 8.4 & 7.7 & 6.6 & 8.6 & 9.6 & 6.5 & 7.2 \\
\midrule
Judgements (count) & 1393 & 1471 & 1537 & 1481 & 1127 & 1057 & 1300 & 1179 \\
\bottomrule
\end{tabular}%
}
\caption{Distribution of the consistency-judge labels on \ACL, per system (Ll.: Llama, Qw.: Qwen, E9/E22: Euro9/Euro22) and prompting condition (percentage of that system's Q2 judgements; last row: absolute counts). The first four labels are coherent (\VarJ), the last three inconsistent (\VarNJ). `register' was never chosen; `facet' and `avoid repetition' are grouped under ``Other coherent''; rare out-of-registry answers are counted with gratuitous divergence.}
\label{tab:q2-labels}
\end{table}

\section{Judgement Counts and Cost}
\label{app:cost}
A \emph{unique judgement} is one judge prompt actually sent to the LLM. Three mechanisms keep this number low: deterministic shortcuts skip cases that need no judge, Q1 propagation lets identical translations of the same source form share one judgement, and a cache serves identical prompts, so repeated runs add no calls. With \texttt{gpt-4.1-mini} at list price, evaluating \biomqmterms\ took about 12{,}000 judgements (about \$6), the eight \iwslt\ outputs about 6{,}500 (about \$7) and the eight \ACL\ outputs about 46{,}000 (about \$50). \STEP\ is the exception: for the validation of \S\ref{ssec:step-protocol} it deliberately runs the exhaustive path with no shortcut, so its 3{,}377 occurrences (one system) yield about 3{,}700 judgements. For comparison, glossary-conformity metrics issue no LLM call, neural QE needs one GPU inference per segment, and \textsc{Gemba-MQM} issued about 4{,}500 calls to \texttt{gpt-4.1-mini} on \biomqmterms, for about \$2. These counts exclude preprocessing, whose alignment step issues one \texttt{gpt-4.1-mini} call per occurrence: about \$5 for \biomqmterms, \$4 for the eight \iwslt\ outputs and \$33 for the eight \ACL\ outputs at list price.

\onecolumn
\ifdefined\nolinenumbers\nolinenumbers\fi
\section{LLM Prompts}
\label{app:prompts}

All prompts are templated with Python \texttt{str.format} placeholders (in braces) populated at runtime. The domain name is injected per corpus through the \texttt{\{domain\}} placeholder, which presents the term under judgement as belonging to that domain (e.g., ``in the biomedical domain''); the context blocks (\texttt{\{concept\_block\}}, \texttt{\{preceding\_segments\}}, \texttt{\{prior\_renderings\}}, \texttt{\{discourse\_context\}}, \texttt{\{concept\_renderings\}}) are built per occurrence, the preceding-segment block holding the five preceding aligned segment pairs. Each judge receives a fixed system message identifying its role, and answers with a JSON object whose \texttt{justification} field precedes the label, so that the label must follow from the stated reasoning. The judges decode at temperature 0. The hapax judgements reuse the Q1 prompt verbatim (\S\ref{ssec:q1}).

\subsection{Q1: Error Judgement}
\label{app:prompt-q1}

\begin{lstlisting}
You are judging how a machine-translation system rendered ONE specialised source term, from English into French in the {domain} domain. The question is ABSOLUTE: is the target a valid, domain-appropriate rendering of this source term? Decide whether it is an ERROR (one label below) or acceptable (no_error).

EVERY alignment is judged here, including those whose target matches the rendering already used in this document. Such a match is not by itself proof of correctness — it may simply repeat the system's own output. Judge the target itself.

Error typology — answer with exactly ONE label. The first seven labels distinguish fine-grained mechanisms of wrong equivalent selection; the following six each group several related mechanisms.
- wrong_related_term        : the target is a term that exists in the {domain} domain but denotes a neighbouring concept, not the one named by the English term (e.g. for "tremor" the MT output "tremblement de terre" names a different concept, the earthquake; expected: "trémor"). Not this label: if the target is not a domain term at all, see general_word_not_term or invented_calque.
- paraphrase_not_term       : the target replaces the established term with a descriptive multi-word paraphrase in general language (e.g. for "lithospheric mantle" the MT output "couche profonde sous la croûte terrestre" is a description, not the term; expected: "manteau lithosphérique").
- invented_calque           : the target is a literal creation of the system that is neither an attested term of the {domain} domain nor a general-language expression in French (e.g. for "earthquake swarm" the MT output "sillage sismique" exists nowhere; expected: "essaim sismique").
- left_untranslated         : the source term was left in English although an established French equivalent exists (e.g. for "megathrusts" the MT output "megathrusts" is untranslated; expected: "mégachevauchements").
- dnt_violated              : a Do-Not-Translate item (proper name, brand, gene or product code, standardised nomenclature symbol, acronym conventionally kept) was translated, or its letters were reordered/reworked (e.g. for "Yellowstone Caldera" the MT output "Caldeira de la Pierre Jaune" translates a proper name; expected: "Caldeira de Yellowstone"; for "MORB" the MT output "BDRM" reworks the acronym; expected: "MORB").
- wrong_component           : one or more constituents of a multi-part term are lexically wrong while the others are correct, yielding a partially correct hybrid term (e.g. for "fault patches" the MT output "taches de faille" mistranslates "patches" while "faille" is right; expected: "zones de faille"). Not this label: if the whole target is a different attested domain term, that is wrong_related_term.
- general_word_not_term     : a single general-language word used instead of the specialised term — unlike paraphrase_not_term, one lexical unit rather than a descriptive phrase (e.g. for "repeaters" the MT output "répétiteurs" is a general-language word with no terminological value; expected: "séismes répétitifs").
- misparsed_structure       : the internal structure of the complex term or noun phrase is mis-analysed — the head noun is misidentified (the head fixes the base category of the concept), a modifier is attached to the wrong constituent, or the constituents are ordered against French conventions (e.g. for "subduction-zone serpentinites" the MT output "zones de subduction constituées de serpentinite" makes the wrong noun the head; expected: "serpentinites en zone de subduction"; for "slow earthquake" the MT output "lent séisme" has the wrong order; expected: "séisme lent").
- wrong_constituent_relation: the semantic relation between the term's constituents (often implicit in English) is rendered wrongly — a wrong linking relation or unfolding, a coordination misread, or a shared (factorised) constituent not distributed to all conjuncts (e.g. for "fast and slow events" the MT output "événements à la fois rapides et lents" misreads two distinct event types as simultaneous properties; expected: "événements rapides et lents"; for "pressure-solution and dislocation creep" the MT output "pression-solution et le glissement par dislocation" fails to distribute "creep" to both conjuncts; expected: "fluage par dissolution-précipitation et par dislocation").
- failed_transposition      : the synthetic English structure is not transposed to the analytic French one — a necessary explicitation (preposition, article, unfolding) is missing, or the English syntax is calqued although the words are attested (e.g. for "strike-slip and subduction thrust faults" the MT output "failles de décrochement et de chevauchement de subduction" lacks the needed explicitation; expected: "failles de décrochement et de chevauchement dans les zones de subduction"; for "SSEs" the MT output "SSEs" erroneously keeps the English plural mark; expected: "SSE", French sigles are invariable).
- grammar_phraseology       : the term choice is right but the grammar or phraseology around it is wrong — determiner or gender/number agreement, a missing or wrong preposition, a non-idiomatic collocation for the domain, or a spelling error (e.g. for "alkaline basalt" the MT output "la basalte alcalin" has the wrong gender; expected: "le basalte alcalin"; for "aseismic slip" the MT output "glissement aseismique" is a misspelling; expected: "glissement asismique"). Not this label: for "aseismic slip" the MT output "glissement asismique" is the correctly spelled attested term — that is no_error.
- content_altered           : the content is altered — an element absent from the source term is added, one or more constituents (or the whole term) are omitted, or the output is unintelligible or unrelated to the source (e.g. for "fault creep" the MT output "glissement lent de la faille" adds "lent", absent from the source; expected: "fluage de faille"; for "earthquake ruptures" the MT output "ruptures d'éruptions" hallucinates an unrelated word; expected: "ruptures sismiques"). If the aligned target is (empty), the term was omitted, so use content_altered — unless the concept is conveyed elsewhere in the French segment, in which case no_error.
- defective_register_form   : the meaning is right but the target-language expression is defective — the weighting or punctuation hierarchy of the constituents is mishandled, or the register is inappropriate for specialised discourse (e.g. for "earthquake ruptures" the MT output "ruptures de tremblements de terre" is correct in meaning but general-public in register; expected: "ruptures sismiques").
- no_error                  : the target conveys the same specialised concept correctly and idiomatically — the exact established term, an attested terminological synonym, a standard abbreviation or acronym, an accepted orthographic variant, a grammatical inflection, or an established domain loanword. Whether a correct-but-different choice is CONSISTENT with the rest of the document is decided in a separate later step, so do NOT judge consistency here.

How to decide (reason in this order, then output exactly one label):
1. Establish the expected French term yourself: the form attested for this concept in specialised {domain} usage (domain corpora, terminology databases, the field's literature). There is not necessarily a single mandatory form. Note that the expected form depends on the SOURCE FORM at hand: an acronym calls for the established acronym, not for the spelled-out head term.
2. Weigh the supporting sections below (concept, preceding segments, prior renderings) for what they are: support, never a verdict.
3. If the target conveys the concept correctly AND is itself an attested, idiomatic form of it, the answer is no_error — whether or not it coincides with the expected form.
4. Otherwise, locate where the defect lies and assign the SINGLE most specific label for it:
   - a wrong terminological equivalent was selected -> wrong_related_term / paraphrase_not_term / invented_calque / left_untranslated / dnt_violated / wrong_component / general_word_not_term;
   - the equivalents are right but the complex term / noun phrase is mis-structured (head, modifier attachment, constituent order) -> misparsed_structure;
   - the constituents are right but their semantic relation is wrong (linking, factorisation) -> wrong_constituent_relation;
   - the English structure is not transposed to French (missing explicitation, syntactic calque) -> failed_transposition;
   - the term is right but its grammar or phraseology in context is wrong (agreement, preposition, collocation, spelling) -> grammar_phraseology;
   - content is added, omitted, or hallucinated -> content_altered;
   - the meaning is right but the target-language expression or register is inappropriate -> defective_register_form.
   When a preposition or an unfolding is involved, separate the three by mechanism: wrong_constituent_relation = the semantic relation between constituents was misread; failed_transposition = the relation is understood but the compact English structure is kept untransposed; grammar_phraseology = the analysis is right and only the grammatical marking (agreement, preposition choice, spelling) is wrong.

{concept_block}{preceding_segments}{prior_renderings}Current segment (the one containing the pair under evaluation):
EN: {src_segment}
FR: {tgt_segment}

Aligned pair under evaluation:
- English term as it appears in the segment: "{src_form_raw}"
- French translation aligned by the system: "{tgt_aligned_raw}"

Return ONLY a JSON object with this exact schema, keys in this order, with no code fences and no text before or after — "justification" is your reasoning, and "label" must follow from it:
{{
  "justification": "<one short sentence: name the expected French form and locate the defect, or state why the target is a valid attested form>",
  "label": "<wrong_related_term|paraphrase_not_term|invented_calque|left_untranslated|dnt_violated|wrong_component|general_word_not_term|misparsed_structure|wrong_constituent_relation|failed_transposition|grammar_phraseology|content_altered|defective_register_form|no_error>"
}}
\end{lstlisting}

\clearpage
\subsection{Q2: Consistency Judgement}
\label{app:prompt-q2}

\begin{lstlisting}
You are judging ONE occurrence of a specialised term in an English-to-French translation in the {domain} domain. The terminology has already been judged CORRECT by a previous step: do NOT re-judge whether the target denotes the right concept, and never report a terminology error here.

Your only question is DOCUMENT CONSISTENCY: given how this concept is rendered elsewhere in this document, is this rendering a coherent choice, or does it introduce an inconsistency? Answer from the document evidence below — no external reference is required, and none may be available.

Three things make a rendering NOT coherent:
1. NEUTRALISATION — the target erases a distinction the author made. The source document uses SEVERAL distinct forms for this concept (a head term and its variants), and this occurrence renders the present source form with the rendering that belongs to ANOTHER source form, collapsing two source distinctions into one French form (e.g. the source uses both "dynamic fault slip" and "dynamic slip", and both end up rendered "glissement dynamique": the distinction between the two source forms is erased). Check each source form's renderings in the evidence below. Exception: if the merged form is exactly the rendering the document has already established for this concept AND the distinction carries no function here, it is coherent — label synonym_merge.
2. SYNONYM INCONSISTENCY — the concept already has an established rendering in this document, and this occurrence switches to a DIFFERENT ADMITTED designation (the evidence line says whether the target is an admitted synonym) without doing any communicative work (e.g. the document has established "fluage de faille" for "fault creep", and this occurrence switches to "fluage de la faille" for no visible reason).
3. GRATUITOUS DIVERGENCE — same switch, but the divergent rendering is NOT an admitted synonym of the concept (it was judged a valid attested form in the previous step, yet it departs from both the document's established usage and the admitted designations) and no communicative work justifies the departure.

A rendering IS coherent when it repeats the document's established form, or when the divergence does genuine communicative work from this closed list — the need must be visible in the current segment itself; do not use these labels as default excuses:
- introduction        : first mention of the concept in the document; no rendering is established yet, so there is nothing to be inconsistent with.
- explicitation       : first mention, and the rendering expands or clarifies the term (e.g. unfolds an acronym, adds the head noun).
- avoid_repetition    : the established rendering already occurs in the SAME sentence, so repeating it would be stylistically heavy. Proximity in the same sentence is required — occurrences in earlier segments do NOT justify this label.
- document_usage      : the target repeats the rendering already established in this document for this source form.
- register            : the current passage itself visibly calls for a different register (e.g. a quoted passage, a glossary definition, an explicitly popularising aside).
- facet               : the divergent rendering highlights a facet of the concept that the current segment explicitly thematises.
- synonym_merge       : the neutralisation exception defined above — in particular when the source forms differ only by a syntactic reordering or trivial variation that French cannot reproduce (e.g. "loading rate" and "rate of loading" can only both be rendered "taux de chargement": the merge is forced by the language, not a system choice).
Length or fluency alone is NOT communicative work.

If this is the FIRST mention of the concept and no rendering is yet established (the evidence line says so), the choice is coherent: label introduction, or explicitation if it expands the term.

How to decide (reason in this order, then output exactly one label):
1. If this is the first mention -> introduction or explicitation. Stop.
2. Compare the target with the renderings already established for THIS source form. If identical to the DOMINANT rendering (marked in the evidence) -> document_usage. Stop. If identical only to a MINORITY prior rendering, this is not self-justifying — an inconsistency repeated is still an inconsistency; continue to steps 3-4.
3. If the target matches the rendering established for ANOTHER source form of this concept -> neutralisation, unless the synonym_merge exception applies; name in the justification the other source form whose rendering this occurrence reproduces.
4. Otherwise the target diverges from the established usage: if the current segment visibly justifies it (closed list above) -> that coherent label; if not, check the target against the French glossary designations in the Concept line: listed -> synonym_inconsistency; not listed -> gratuitous_divergence. If the Concept line says the concept has no glossary entry, the admitted/gratuitous distinction cannot be made: use synonym_inconsistency — the previous step already validated the rendering as an attested designation of this concept.

CAUTION: the document renderings below are THIS SAME SYSTEM's earlier outputs. They show what the document has established in practice — which is exactly what consistency is measured against — but they are not proof of correct French.

{discourse_context}{concept_renderings}Current segment (the one containing the occurrence under evaluation):
EN: {src_segment}
FR: {tgt_segment}

Occurrence under evaluation:
- English source form as it appears: "{src_form_raw}"
- Is this source form the concept's head term or a variant? {head_or_variant}
- French translation aligned by the system: "{tgt_aligned_raw}"
- {concept_glossary_line}
- Is the target one of the French glossary designations above? {is_glossary_designation}
- Is this the first mention of the concept in the document? {is_first_mention}

Return ONLY a JSON object with this exact schema, keys in this order, with no code fences and no text before or after — "justification" is your reasoning, and "label" must follow from it:
{{
  "justification": "<one short sentence: name the document's established rendering and state why this occurrence is coherent with it or diverges from it; for neutralisation, name the other source form whose rendering is reproduced>",
  "label": "<introduction|explicitation|avoid_repetition|document_usage|register|facet|synonym_merge|neutralisation|synonym_inconsistency|gratuitous_divergence>"
}}
\end{lstlisting}

\clearpage
\subsection{Reference Selection}
\label{app:prompt-select}

Used by the \texttt{validated} cascade when an LLM must select among observed candidate translations (\S\ref{ssec:reference-selection}).

\begin{lstlisting}
You are given one English specialised source term in the {domain} domain and the French translations observed for it across several machine-translation systems in a single document. Pick the best reference translation: the single French form a professional translator of this domain would use consistently for this term in this document. It becomes the anchor against which every system's translation of this term is judged.

Selection rules, in order:
1. Prefer the glossary candidate (the line tagged "glossary:") when there is one, unless it clearly renders a different sense of the term than the sense used in this document. A glossary candidate is shown only for a head term; a variant source form has none.
2. Otherwise prefer the most idiomatic, terminologically appropriate French form. An established concise term beats a longer explanatory paraphrase; never prefer a form merely because it is longer.
3. Tie-break: higher observed count; if still tied, the one appearing first (candidate list first, then observed translations top to bottom).

The chosen reference MUST be one of the French forms displayed below. Never invent, correct, translate back, or rephrase a form. Copy it exactly, WITHOUT the surrounding quotes and WITHOUT the strategy tag.

In the two lists below, each form is wrapped in quotes; a candidate line is additionally prefixed by a strategy tag ("glossary", "frequency" or "first"). The quotes and the tag are display formatting and must not be copied into your answer; the tag only tells you where a candidate came from (Rule 1 uses it to spot the glossary candidate). A prefLabel shown as "(unknown)", or a list shown as "(none)", is unavailable.

<examples>
Illustrative of the decision types (they may be from a different domain than the item you decide). Each example shows the same fields as the item you will decide. Base each decision only on the fields shown.

<example>
Term and document:
- English source term: seismic hazard
- Glossary concept (English prefLabel): seismic hazard
- Document: doc3
Candidate references (each line: "- <strategy>: <French form in quotes>"):
  - glossary: "aléa sismique"
  - frequency: "risque sismique"
  - first: "risque sismique"
Observed French translations in this document (each line: "- <French form in quotes> × <count>"):
  - "risque sismique" × 2
  - "aléa sismique" × 1
{{
  "reference": "aléa sismique",
  "justification": "Rule 1 prefers the glossary candidate 'aléa sismique', the domain-standard term, over the more frequent 'risque sismique'."
}}
</example>

<example>
Term and document:
- English source term: stress
- Glossary concept (English prefLabel): stress
- Document: doc9
Candidate references (each line: "- <strategy>: <French form in quotes>"):
  - glossary: "stress"
  - frequency: "contrainte"
  - first: "contrainte"
Observed French translations in this document (each line: "- <French form in quotes> × <count>"):
  - "contrainte" × 4
  - "stress" × 1
{{
  "reference": "contrainte",
  "justification": "Rule 1 exception: the glossary form 'stress' fits the psychological sense, but this document uses the mechanical sense, whose established term is 'contrainte'."
}}
</example>

<example>
Term and document:
- English source term: crust-mantle boundary
- Glossary concept (English prefLabel): Moho
- Document: doc5
Candidate references (each line: "- <strategy>: <French form in quotes>"):
  - frequency: "limite entre la croûte et le manteau"
  - first: "limite entre la croûte et le manteau"
Observed French translations in this document (each line: "- <French form in quotes> × <count>"):
  - "limite entre la croûte et le manteau" × 3
  - "limite croûte-manteau" × 2
{{
  "reference": "limite croûte-manteau",
  "justification": "No glossary candidate; rule 2 prefers the concise compound term 'limite croûte-manteau' over the longer paraphrase, even though the paraphrase is more frequent."
}}
</example>

<example>
Term and document:
- English source term: very-low-frequency earthquakes
- Glossary concept (English prefLabel): (unknown)
- Document: doc1
Candidate references (each line: "- <strategy>: <French form in quotes>"):
  - frequency: "séismes de très basse fréquence"
  - first: "séismes de très basse fréquence"
Observed French translations in this document (each line: "- <French form in quotes> × <count>"):
  - "séismes de très basse fréquence" × 2
  - "séismes à très basse fréquence" × 2
{{
  "reference": "séismes de très basse fréquence",
  "justification": "'de' and 'à' are both attested and equally idiomatic here and the two forms tie on count, so rule 3 keeps the one appearing first."
}}
</example>

<example>
Term and document:
- English source term: SEM
- Glossary concept (English prefLabel): scanning electron microscope
- Document: doc8
Candidate references (each line: "- <strategy>: <French form in quotes>"):
  - frequency: "microscopie électronique à balayage"
  - first: "microscopie électronique à balayage"
Observed French translations in this document (each line: "- <French form in quotes> × <count>"):
  - "microscopie électronique à balayage" × 2
  - "MEB" × 1
{{
  "reference": "MEB",
  "justification": "No glossary candidate; rule 2 prefers the established concise acronym 'MEB' over its longer spelled-out form, even though the spelled-out form is more frequent."
}}
</example>
</examples>

Term and document:
- English source term: {src_form}
- Glossary concept (English prefLabel): {preflabel_en}
- Document: {doc_id}
Candidate references (each line: "- <strategy>: <French form in quotes>"):
{candidates}
Observed French translations in this document (each line: "- <French form in quotes> × <count>"):
{distribution}

Return ONLY a JSON object with this exact schema:
{{
  "reference": "<the chosen French form, copied exactly, no surrounding quotes, no strategy tag>",
  "justification": "<one short sentence giving the decisive reason>"
}}
\end{lstlisting}

\clearpage
\subsection{Reference Generation}
\label{app:prompt-generate}

Used by the \texttt{validated} cascade when a vetted heuristic reference is rejected and a reference must be produced from source-side evidence only (\S\ref{ssec:reference-selection}).

\begin{lstlisting}
You are given a specialised concept in the {domain} domain and ONE English source term that designates it. Produce the French term expected for THIS source term: the term a professional translator of this domain would use for it, consistently, in a document. It must be an ESTABLISHED French term, not a plausible invention; it will serve as a candidate reference against which MT systems are judged.

# Source variation category

The "Source variation category" tells you how the source term relates to the concept's HEAD TERM: it is the variation that turns the head term into this source form (both in English). If the source term IS the head term, the category is "none". Treat it as an AID to find the right French reference, not as a constraint: the reference need not mirror the same kind of variation in French. It hints at what the source term is (an acronym, a reduction, a synonym...) so you produce the French form actually established for it; established usage always decides.

- none : the source term is the head term itself.
- graphical variation : same content, different written form — acronym or abbreviation (World Health Organization / WHO), symbol (lead / Pb), spelling or hyphenation (trademark / trade mark).
- reduction : a shortened form — element deleted (airborne dust particle / airborne dust) or an extension dropped (post-authorisation safety study / post-authorisation study).
- expansion : elements added — an explicit feature, a lexical insertion, or the spelled-out form of an abbreviation (redox / oxidation-reduction).
- lexical variation : a different lexical designation of the same concept — synonym, quasi-synonym, or attested translation (stress / constraint).
- morphosyntactic variation : structure or inflection change — constituent order, article, inflection, derivation (employment contract / contract of employment).
- composite form : several of the above combined; apply the reasoning of each component change.

# Rules

1. Let the variation category guide you toward the French designation actually established for this source term. Often the established French form is of the same kind (an acronym stays an acronym, a reduction stays a reduction), but not always: established usage decides, never a mechanical derivation from the category.

2. Acronyms and abbreviations (category "graphical variation", or "expansion" when the source form is the spelled-out expansion of one). The correct French form depends on how the HEAD TERM is translated into French. Decide among three moves:
   (a) Invert the acronym — when the concept is translated into French and that French term has an established French acronym, give the FRENCH acronym. E.g. WHO -> OMS (Organisation mondiale de la santé); emergency contraception (EC) -> contraception d'urgence (CU); DNA -> ADN.
   (b) Keep the source acronym as-is — when the field uses the English acronym unchanged in French. E.g. PCR, BERT, GPT, LLM.
   (c) Spell it out (rare) — when neither a French acronym nor the borrowed source acronym is used, give the developed French form. E.g. ASAP -> dès que possible.
   Do NOT assume French spells the term out by default. Your "reference" must match your reasoning: if you conclude the source acronym is used as-is, return that acronym.

3. Keep the reference's SCOPE identical to the source term: add no word it does not contain, drop none it does.

4. For a lexical variation, give the established French term proper to THAT synonym, not the French head term.

5. If French keeps a non-acronym source form unchanged (a proper name, a do-not-translate item), return it unchanged.

6. Do NOT invent. For a non-acronym form, if no established French designation exists and you are not confident, set "generatable" to false. For an acronym, keep the source acronym (rule 2b) instead of declining.

7. Return the reference in canonical form: lemma (singular unless the term is inherently plural), lowercase except proper nouns and acronyms, no surrounding quotes.

# Examples

Each example shows the same fields as the Data block below, then the expected JSON. The examples are illustrative only: NEVER assume the current term behaves like an example it superficially resembles; established usage for THIS term decides.

Example 1
- English source form: WHO
- Source variation category: graphical variation
- Concept head term (English): World Health Organization
- Concept established French term (head term), if known: Organisation mondiale de la santé
=> {{"justification": "Acronym inverted: the concept is translated and has an established French acronym.", "generatable": true, "reference": "OMS", "confidence": "high"}}

Example 2
- English source form: PCR
- Source variation category: graphical variation
- Concept head term (English): polymerase chain reaction
- Concept established French term (head term), if known: réaction en chaîne par polymérase
=> {{"justification": "Acronym kept as-is: standard French usage borrows PCR unchanged.", "generatable": true, "reference": "PCR", "confidence": "high"}}

Example 3
- English source form: oxidation-reduction
- Source variation category: expansion
- Concept head term (English): redox
- Concept established French term (head term), if known: redox
=> {{"justification": "Expansion of an abbreviation: the spelled-out French form of « redox ».", "generatable": true, "reference": "oxydoréduction", "confidence": "high"}}

Example 4
- English source form: abiotic constraint
- Source variation category: lexical variation
- Concept head term (English): abiotic stress
- Concept established French term (head term), if known: stress abiotique
=> {{"justification": "Lexical variation: French equivalent of the synonym, not of the head term.", "generatable": true, "reference": "contrainte abiotique", "confidence": "medium"}}

Example 5
- English source form: post-authorisation study
- Source variation category: reduction
- Concept head term (English): post-authorisation safety study
- Concept established French term (head term), if known: étude de sécurité post-autorisation
=> {{"justification": "Reduction: the French head term with the dropped element ('sécurité') removed.", "generatable": true, "reference": "étude post-autorisation", "confidence": "medium"}}

# Data

- English source form: {src_form}
- Source variation category: {variation_category}
- Concept head term (English): {preflabel_en}
- Concept established French term (head term), if known: {preflabel_fr}
English sentences where the source form occurs (evidence of usage, may mention other variants of the concept; possibly truncated):
{contexts}

Return ONLY a JSON object with this exact schema, keys in this order, with no code fences and no text before or after — "justification" is your reasoning; "generatable", "reference" and "confidence" must follow from it:
{{
  "justification": "<one short sentence: how you derived the form, or why no established form exists>",
  "generatable": <true or false, bare JSON boolean>,
  "reference": "<the canonical French form; empty string if generatable is false>",
  "confidence": "<high|medium|low — your confidence that this exact form is the established one>"
}}
\end{lstlisting}

\clearpage
\subsection{Term Alignment}
\label{app:prompt-align}

Used by the preprocessing to align each detected occurrence with its translation in the target segment (Appendix~\ref{app:pipeline}), after the system message ``You are a professional translator specializing in terminology alignment.'' The instructions are those of the organisers of the WMT25 terminology task \citep{semenov-etal-2025-findings}; the six in-context examples are ours.

\begin{lstlisting}
You are a professional English-French translator, teaching the students the course on technical translation. You are checking a student's translation of a sentence that contains a technical term. You are given an English term (it can be a word or an expression), a source English sentence containing this term (it may be cased differently or contain additional punctuation), and a student's French translation. You need to find how the student has translated the term in question in French, and return only that term.

Important: do not change the translated term anyhow, copy it straight from the sentence! For example, keep the casing and the grammar form of the translated term as is.

When completing the task, follow the examples below:

English sentence: This paradigm based on ant colony algorithms for the exploration of the graph removes the need to dynamically expand the graph: the memory footprint becomes independant of the language model size.
English term: language model
French translation: Ce paradigme basé sur les algorithmes de colonie de fourmis pour l'exploration du graphe supprime le besoin d'étendre dynamiquement le graphe : l'empreinte mémoire devient indépendante de la taille du modèle de langage.
Translated term: modèle de langage

English sentence: In this article we presented a new paradigm to expand word graphs in automatic speech recognition systems.
English term: automatic speech recognition systems
French translation: Dans cet article, nous avons présenté un nouveau paradigme pour l'expansion des graphes de mots dans les systèmes de reconnaissance automatique de la parole.
Translated term: systèmes de reconnaissance automatique de la parole

English sentence: Evolution of rescoring according the number of ant by node (on the dev set) during 1 run and comparison between Viterbi beam-search and ant colonies algorithm: Computing time against performance on the dev set with a 4-gram language model.
English term: dev set
French translation:  Évolution du ré-évaluation en fonction du nombre d’ant par nœud (sur l’ensemble de développement) au cours d’une exécution et comparaison entre la recherche en faisceau Viterbi et l’algorithme des colonies de fourmis : temps de calcul par rapport aux performances sur l’ensemble de développement avec un modèle de langage à 4 grammes.
Translated term: ensemble de développement

English sentence:   In Earth sciences, plate tectonics took a very long time to be accepted in the 1960s, and the origin of global warming was still the subject of heated debate very recently.
English term: plate tectonics
French translation: En sciences de la Terre, la théorie de la tectonique des plaques a mis très longtemps à être acceptée dans les années 1960, et l'origine du réchauffement climatique était encore très récemment l'objet de débats houleux.
Translated term: tectonique des plaques

English sentence: The first SR15 Special Report (October 2018) focuses on the impacts associated with a global warming of 1.5°C, as well as compatible greenhouse gas emission trajectories, in the context of strengthening the response to climate change, sustainable development and efforts to eradicate poverty:
English term: greenhouse gas emission
French translation: Le premier rapport spécial SR15 (octobre 2018) se concentre sur les impacts associés à un réchauffement planétaire de 1,5°C, ainsi que sur les trajectoires d’émissions de gaz à effet de serre compatibles, dans le contexte du renforcement de la lutte contre le changement climatique, du développement durable et des efforts d’éradication de la pauvreté :
Translated term: émissions de gaz à effet de serre

English sentence: We used our set of manually-written chronologies as a training corpus to perform machine learning experiments.
English term: training corpus
French translation: Nous avons utilisé notre ensemble de chronologies rédigées manuellement comme corpus d'apprentissage pour effectuer des expériences d'apprentissage automatique.
Translated term: corpus d'apprentissage

English sentence: {src_segment}
English term: {src_term}
French translation: {tgt_segment}
Translated term:
\end{lstlisting}

\clearpage
\subsection{Variation Labelling}
\label{app:prompt-variation}

Used by the preprocessing to assign variation categories (Appendix~\ref{app:pipeline}), with \texttt{gpt-4.1} at temperature 0, after the system message ``You are an expert in scientific terminology and linguistic variation analysis.'' It is a revised version of the prompt of \citet{dahan-etal-2026-improving}. The \texttt{\{context\_section\}} block lists up to three corpus sentences containing the head term and up to three containing the variant, or notes that one of them is not found in the corpus.

\begin{lstlisting}
You are an expert in scientific terminology and linguistic variation, teaching a course on terminology. A student has identified a variant of a head term and needs your help to classify the relationship between the head term and its variant according to the established typology.

You receive:
- A head term (the preferred form of a technical term)
- A variant (an alternative term that qualifies the head term)
- Context examples showing how these terms appear in real scientific texts (when available)

Your task: Determine the type of variation between the head term and its variant. Classify according to the variation typology. Return ONLY a JSON object with the exact format shown in the examples below.

Reference of variation types (in the following example for each label we write headterm/variant ):

VG (Graphical): Written form changes without semantic change (acronyms, spelling, symbols)
  → VG1: Acronym/initialism (European Parliament/EP, non-governmental organisation/NGO, coronavirus disease 2019/COVID-19)
  → VG2: Symbol/formula (lead/Pb, CO₂)
  → VG3: Spelling change (hyphen, space, case, regional: trade mark/trademark, penalise/penalize, case law/case-law)
  → VG4: Multiple spelling changes (coronavirus disease 2019/Coronavirus Disease-2019)
  → VG5: Partial abbreviation of one constituent (apparent polar wander path/APW path)

VR (Reduction): Deletion of one or more constituents
  → VR1: Base reduction — deletion of a base (automatic translation system/automatic translation)
  → VR2: Extension reduction — deletion of qualifier (coronavirus disease 2019/coronavirus disease, post-authorisation safety study/post-authorisation study)
  → VR3: Other reduction — ellipsis, blending, or non-standard truncation (middle-capitalisation company/mid-cap)

VE (Expansion): Addition of one or more constituents
  → VE1: Semantic addition — relevant but redundant attribute (coronavirus disease 2019/pandemic coronavirus disease 2019)
  → VE2: Explicit form — substitution of a morpheme/lexeme by a fuller form (uninterrupted/without interruption)
  → VE3: Lexical insertion — nominal, verbal, or paraphrastic addition (dark urine/dark-colored urine)
  → VE4: Abbreviation development — expansion of a short form into simple or complex unit (redox/oxidation-reduction)

VL (Lexical): Lexical substitution (synonyms, near-synonyms, translations validated as variant in context)
  → VL1: Simple unit substitution (residues/waste)
  → VL2: Complex unit — base change only (coronavirus disease 2019/coronavirus infection 2019, action for damages/claim for damages)
  → VL3: Complex unit — extension/attribute change only (coronavirus disease 2019/novel coronavirus disease, maritime law/marine law, action for damages/action for compensation)
  → VL4: Complex unit — base AND extension change (coronavirus disease 2019/Wuhan pneumonia, action for damages/claim for compensation)

VMS (Morphosyntactic): Structure or inflection changes (word order, articles, inflection, derivation)
  → VMS1: Constituent order change (coronavirus disease 2019/2019 coronavirus disease)
  → VMS2: Addition/removal of article (fixation of nitrogen/fixation of the nitrogen)
  → VMS3: Inflection change (buying cartel/buyer's cartel, law of tort/law of torts, pass-on of overcharges/passing-on of overcharges)
  → VMS4: Morphological change (statutory law/statute law, unmargined derivative/non-margined derivative)
  → VMS5: Structural change — prepositional, adjectival, or syntactic reanalysis (contract of employment/employment contract, asylum application/application for asylum, misconduct in office/official misconduct)

CM (Multiple Changes): Combination of two or more of the above types — combine codes with + (e.g. VG2+VL1, VR1+VMS5)

Important rules:
1) The (head term, variant) pair you receive is ALWAYS a true terminological
   variation. Your task is to identify which type of variation it is.
2) Count the TYPES of changes, not the number of words: one type → VG/VMS/VR/VE/VL,
   two or more types → CM.
3) **Acronym rule (critical)**: A variant that looks like an acronym is VG1 ALONE
   *only* if every letter of the acronym maps to a word actually present in the
   head term. If the acronym contains one or more letters referring to words
   ABSENT from the head term, the variant carries an implicit semantic addition
   and must be classified as CM = VG1+VE1 (or VG1+VL1/VL4 if a word is replaced
   rather than added). Examples of this hidden-CM pattern:
     - "topic detection" / "TDT": the final T refers to "tracking" (absent) → CM VG1+VE1
     - "phrase" / "PP": "P" refers to "prepositional" (absent) → CM VG1+VE1
     - "language model" / "LM": only 2 letters but the head term itself is fine — this IS VG1.
   Always check letter-by-letter coverage before settling on plain VG1.
4) **Partial abbreviation rule (VG5 vs VG1)**: If part of the head term is preserved
   verbatim in the variant (e.g. "automatic speech recognition system" / "ASR system"
   keeps "system"), this is VG5 (partial abbreviation), NOT VG1 (full acronym).
5) **Complex unit rule (VL2/VL3 vs VL1)**: VL1 is reserved for substitution of a
   *simple* (1-word) head term. If the head term has ≥ 2 words and only the base
   (or only the modifier) is replaced, classify as VL2 / VL3 — NEVER VL1.
6) For CM, combine codes with + (e.g., "VG1+VE1", "VL2+VR2", "VMS1+VR2+VL3").
7) Copy the exact JSON format from the examples — no markdown, no extra text.
8) Keep justifications brief (1-2 sentences maximum).

When completing the task, follow the examples below:

Head term: European Parliament
Variant: EP
{"category": "VG", "subtype": "VG1", "justification": "Acronym formed from the initials of the head term."}

Head term: carbon dioxide
Variant: CO₂
{"category": "VG", "subtype": "VG2", "justification": "Chemical formula symbol replacing the complete term."}

Head term: trade mark
Variant: trademark
{"category": "VG", "subtype": "VG3", "justification": "Orthographic change: removal of space between words."}

Head term: coronavirus disease 2019
Variant: Coronavirus Disease-2019
{"category": "VG", "subtype": "VG4", "justification": "Multiple orthographic changes: capitalization and hyphen added."}

Head term: apparent polar wander path 
Variant: APW path
{"category": "VG", "subtype": "VG5", "justification": "Partial abbreviation of the head term."}

Head term: automatic translation system
Variant: automatic translation
{"category": "VR", "subtype": "VR1", "justification": "Base reduction: deletion of the base 'system'."}

Head term: coronavirus disease 2019
Variant: coronavirus disease
{"category": "VR", "subtype": "VR2", "justification": "Extension reduction: deletion of the modifier '2019'."}

Head term: middle-capitalisation company
Variant: mid-cap
{"category": "VR", "subtype": "VR3", "justification": "Partial reduction: deletion of the specification 'company'."}

Head term: coronavirus disease
Variant: pandemic coronavirus disease
{"category": "VE", "subtype": "VE1", "justification": "Addition of semantic feature: 'pandemic' specifies the scope."}

Head term: phyllosilicate
Variant: sheet silicate
{"category": "VE", "subtype": "VE2", "justification": "Substitution with explicit descriptive form."}

Head term: dark urine
Variant: dark-colored urine
{"category": "VE", "subtype": "VE3", "justification": "Lexical insertion: addition of '-colored'."}

Head term: redox
Variant: oxidation-reduction
{"category": "VE", "subtype": "VE4", "justification": "Development of abbreviated form into complete form."}

Head term: myocardial infarction
Variant: heart attack
{"category": "VL", "subtype": "VL1", "justification": "Lexical substitution of a simple unit by a near-synonym."}

Head term: stress
Variant: constraint
{"category": "VL", "subtype": "VL1", "justification": "Lexical substitution by translation: borrowed term replaced by native equivalent."}

Head term: translation system
Variant: translation model
{"category": "VL", "subtype": "VL2", "justification": "Base change: 'system' replaced by 'model'."}

Head term: binary classification model
Variant: logistic regression model
{"category": "VL", "subtype": "VL3", "justification": "Extension change: modification of the model type."}

Head term: coronavirus disease 2019
Variant: Wuhan pneumonia
{"category": "VL", "subtype": "VL4", "justification": "Base + extension change: complete lexical substitution."}

Head term: coronavirus disease 2019
Variant: 2019 coronavirus disease
{"category": "VMS", "subtype": "VMS1", "justification": "Change in constituent order: '2019' moved to the beginning."}

Head term: fixation of nitrogen
Variant: fixation of the nitrogen
{"category": "VMS", "subtype": "VMS2", "justification": "Addition of definite article 'the'."}

Head term: law of tort
Variant: law of torts
{"category": "VMS", "subtype": "VMS3", "justification": "Inflectional change from singular to plural."}

Head term: statutory law
Variant: statute law
{"category": "VMS", "subtype": "VMS4", "justification": "Derivation: 'statutory' is the adjectival form of the noun 'statute'."}

Head term: contract of employment
Variant: employment contract
{"category": "VMS", "subtype": "VMS5", "justification": "Structural reanalysis: noun+prep phrase replaced by attributive compound."}

Head term: alpha particle
Variant: α ray
{"category": "CM", "subtype": "VG2+VL1", "justification": "Two changes: 'alpha'→'α' is graphical (VG2), 'particle'→'ray' is lexical (VL1)."}

Head term: automatic translation system
Variant: translation model
{"category": "CM", "subtype": "VL2+VR2", "justification": "Lexical change 'system'→'model' (VL2) and reduction of the extension 'automatic' (VR2)."}

Head term: topic detection
Variant: TDT
{"category": "CM", "subtype": "VG1+VE1", "justification": "Acronym (VG1) but the final 'T' refers to 'tracking', a word absent from the head term, so a semantic feature is implicitly added (VE1)."}

Head term: automatic speech recognition system
Variant: ASR system
{"category": "VG", "subtype": "VG5", "justification": "'system' is preserved verbatim; only 'automatic speech recognition' is abbreviated. This is partial abbreviation (VG5), not a full acronym (VG1)."}

Head term: similarity measure
Variant: similarity function
{"category": "VL", "subtype": "VL2", "justification": "Complex (≥2-word) head term: the base 'measure' is replaced by 'function' while the modifier 'similarity' is preserved → VL2 (base change), not VL1."}

{context_section}
Head term: {head_term}
Variant: {variant}
\end{lstlisting}

\twocolumn
\ifdefined\linenumbers\linenumbers\fi

\end{document}